\documentclass{article}

\usepackage{microtype}
\usepackage{graphicx}
\usepackage{subcaption} 
\usepackage{booktabs} % for professional tables

\usepackage{hyperref}

\usepackage[shortlabels]{enumitem}
\setitemize{leftmargin=*,noitemsep,topsep=0pt,parsep=0pt,partopsep=0pt}
\setenumerate{leftmargin=*,noitemsep,topsep=0pt,parsep=0pt,partopsep=0pt}

\usepackage{bm}

\usepackage{amsmath,amssymb,amsthm}
\usepackage{multirow}
\usepackage{makecell}
\usepackage{cleveref}
\usepackage{todonotes}
\usepackage{natbib}

\newtheorem{theorem}{Theorem}[section]
\newtheorem{definition}{Definition}

\newtheorem{assumption}{Assumption}[section]

\newtheorem{lemma}{Lemma}[section]
\newtheorem{corollary}{Corollary}[section]

\newcommand{\bb}[1]{\mathbb{#1}}
\newcommand{\bv}[1]{\boldsymbol{#1}}
\newcommand{\bvec}[1]{\boldsymbol{#1}}
\newcommand{\ca}[1]{\mathcal{#1}}

\newcommand{\GE}{\text{GE}}

\newcommand{\longversion}[1]{}
\DeclareMathOperator*{\argmax}{arg\,max}

\usepackage[accepted]{icml2020}

\begin{document}

\twocolumn[
\icmltitle{Towards Understanding the Regularization of Adversarial Robustness on Neural Networks}

% It is OKAY to include author information, even for blind
% submissions: the style file will automatically remove it for you
% unless you've provided the [accepted] option to the icml2020
% package.

% List of affiliations: The first argument should be a (short)
% identifier you will use later to specify author affiliations
% Academic affiliations should list Department, University, City, Region, Country
% Industry affiliations should list Company, City, Region, Country

% You can specify symbols, otherwise they are numbered in order.
% Ideally, you should not use this facility. Affiliations will be numbered
% in order of appearance and this is the preferred way.
\icmlsetsymbol{equal}{*}

\begin{icmlauthorlist}
\icmlauthor{Yuxin Wen}{equal,scut,pazhou}
\icmlauthor{Shuai Li}{equal,scut}
\icmlauthor{Kui Jia}{scut,pazhou}
\end{icmlauthorlist}

\icmlaffiliation{scut}{School of Electronic and Information Engineering, South China University of Technology, Guangzhou, Guangdong 510640, China}
\icmlaffiliation{pazhou}{Pazhou Lab, Guangzhou, 510335, China}

\icmlcorrespondingauthor{Shuai Li}{lishuai918@gmail.com}
\icmlcorrespondingauthor{Yuxin Wen}{wen.yuxin@mail.scut.edu.cn}
\icmlcorrespondingauthor{Kui Jia}{kuijia@scut.edu.cn}

% You may provide any keywords that you
% find helpful for describing your paper; these are used to populate
% the "keywords" metadata in the PDF but will not be shown in the document
\icmlkeywords{Adversarial Training, Adversarial Robustness, Statistical
  Learning, Generalization, Regularization, Neural Networks}

\vskip 0.3in
]

% this must go after the closing bracket ] following \twocolumn[ ...

% This command actually creates the footnote in the first column
% listing the affiliations and the copyright notice.
% The command takes one argument, which is text to display at the start of the footnote.
% The \icmlEqualContribution command is standard text for equal contribution.
% Remove it (just {}) if you do not need this facility.

%\printAffiliationsAndNotice{}  % leave blank if no need to mention equal contribution
\printAffiliationsAndNotice{\icmlEqualContribution} % otherwise use the standard text.

\begin{abstract}
The problem of adversarial examples has shown that modern Neural Network (NN) models could be rather fragile. 
Among the more established techniques to solve the problem, one is to require the model to be {\it $\epsilon$-adversarially robust} (AR); that is, to require the model not to change predicted labels when any given input examples are perturbed within a certain range. 
However, it is observed that such methods would lead to standard performance degradation, i.e., the degradation on natural examples.  
In this work, we study the degradation through the regularization perspective.
We identify quantities from generalization analysis of NNs; with the identified quantities we empirically find that AR is achieved by regularizing/biasing NNs towards less confident solutions by making the changes in the feature space (induced by changes in the instance space) of most layers smoother uniformly in all directions; so to a certain extent, it prevents sudden change in prediction w.r.t.  perturbations. 
However, the end result of such smoothing concentrates samples around decision boundaries, resulting in less confident solutions, and leads to worse standard performance. 
Our studies suggest that one might consider ways that build AR into NNs in a gentler way to avoid the problematic regularization.
\end{abstract}

\section{Introduction}
\label{sec:introduction}
Despite the remarkable performance \citep{Krizhevsky2012} of Deep Neural
Networks (NNs), they are found to be rather fragile and easily fooled by
adversarial examples \citep{Szegedy2013}.  More intriguingly, these adversarial
examples are generated by adding imperceptible noise to normal examples, and
thus are indistinguishable for humans.  NNs that are more robust to adversarial
examples tend to have lower standard accuracy \citep{Su}, i.e., the accuracy
measured on natural examples.  The trade-off between robustness and accuracy
has been empirically observed in many works
\citep{DBLP:journals/ml/FawziFF18,Kurakin2017,Madry2017,Turner2018}, and has been theoretically analyzed under the context of simple models,
e.g., linear models \cite{Turner2018}, quadratic models
\cite{DBLP:journals/ml/FawziFF18}, but it is not clear whether the analysis
generalizes to NNs. For example, \citet{Turner2018} show that for linear
models, if examples are close to decision boundaries, robustness provably
conflicts with accuracy, though the proof seems unlikely to generalize to NNs.
Arguably, the most widely used remedy is developed to require NNs to be {\it
$\epsilon$-adversarially robust} (AR), e.g., via Adversarial Training
\citep{Madry2017}, Lipschitz-Margin Training \citep{Tsuzuku}; that is, they
require the model not to change predicted labels when any given input examples
are perturbed within a certain range. In practice, such AR methods are found to
lead to worse performance measured in standard classification accuracy.
Alternatives to build AR into NNs are also being developed. For instance,
\citet{Zhang2019} show that a gap exists between surrogate risk gap and 0-1
risk gap if many examples are close to decision boundaries, and better
robustness can be achieved by pushing examples away from decision boundaries.
But pushing examples away again degrades NN performance in their
experiments. But they are yet to be widely adopted by the community.

We investigate how adversarial robustness built into NNs by the arguably most
established method, i.e., Adversarial Training \citep{Madry2017}, influences
the behaviors of NNs to make them more robust but have lower performance
through the lens of regularization. In an earlier time \citep{Szegedy2013},
adversarial training has been suggested as a form of regularization: it
augments the training of NNs with adversarial examples, and thus might improve
the generalization of the end models.  Note that such a {\it hard requirement}
that the adversarial examples need to be classified correctly is different from
the methods that increase adversarial robustness by adding a soft penalty term to
the risk function employed by \citet{Lyu2016} and \citet{Miyato2018a}, or a
penalty term through curvature reduction \cite{Moosavi-Dezfooli2019}, or local
linearization \cite{Qin2019} (more discussion in
\cref{sec:furth-relat-works}). In these works, regularization is explicitly
enforced by a penalty term, while in adversarial training, it is not clear that
how training with augmented adversarial examples regularizes NNs. For example,
if adversarial training does work as a regularizer, how does a possible
improvement in generalization by using more data end up degrading performance?
Even such a basic problem does not have a clear answer.  To understand the
regularization effects of AR on NNs, we go beyond simple linear or quadratic
models and undertake a comprehensive generalization analysis of AR by
establishing a rigorous generalization bound on NNs, and carrying out a series
of empirical studies theoretically guided by the bound.

Technically, improved generalization implies the reduction in gap between
training errors and test errors.  Regularization achieves the gap reduction by
reducing the size of the hypothesis space, which reduces the variance, but
meanwhile increases the bias of prediction made — a constant classifier can
have zero generalization errors, but also have low test performance.  Thus,
when a hypothesis space is improperly reduced, another possible outcome is
biased poorly performing models with reduced generalization gaps.

\begin{figure*}[t]
    \centering
    \begin{subfigure}[b]{0.41\textwidth}
      \includegraphics[width=\columnwidth]{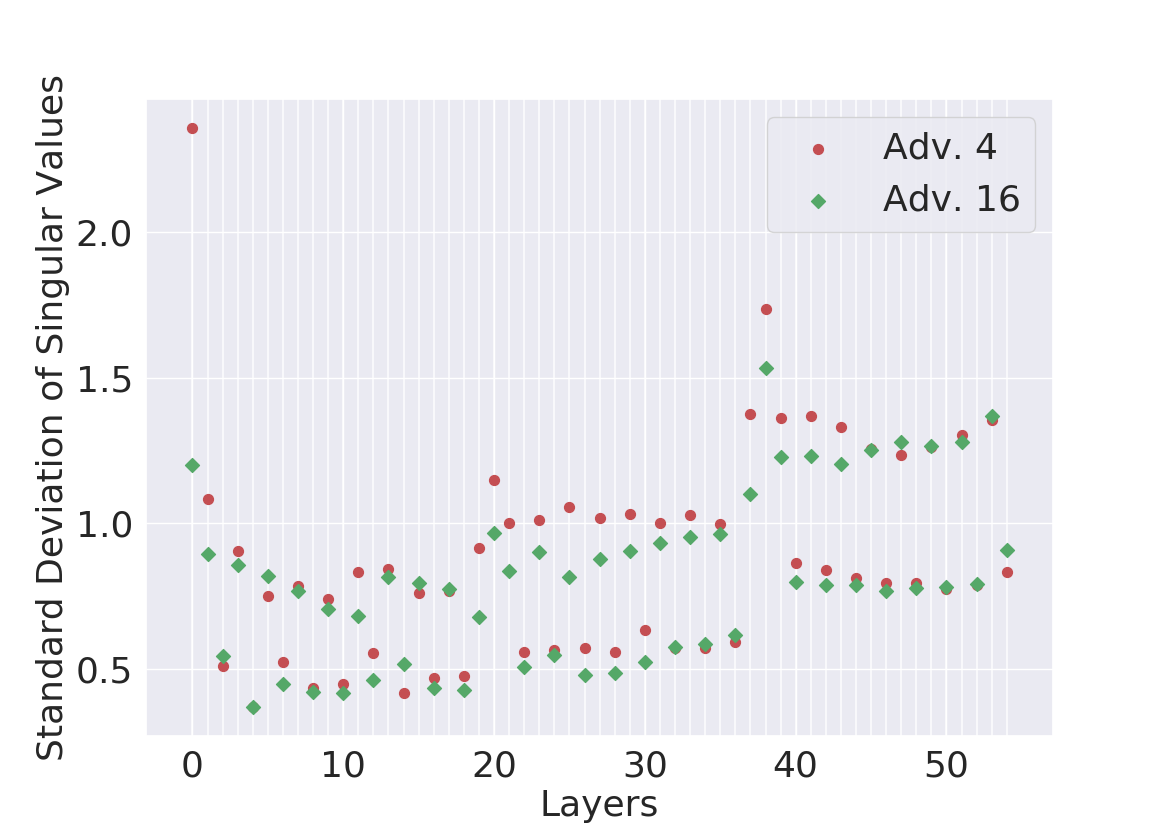}
      \caption{STD of Singular Values}
      \label{fig:singular_value_CIFAR10_pre}
    \end{subfigure}
    \begin{subfigure}[b]{0.41\textwidth}
      \includegraphics[width=\columnwidth]{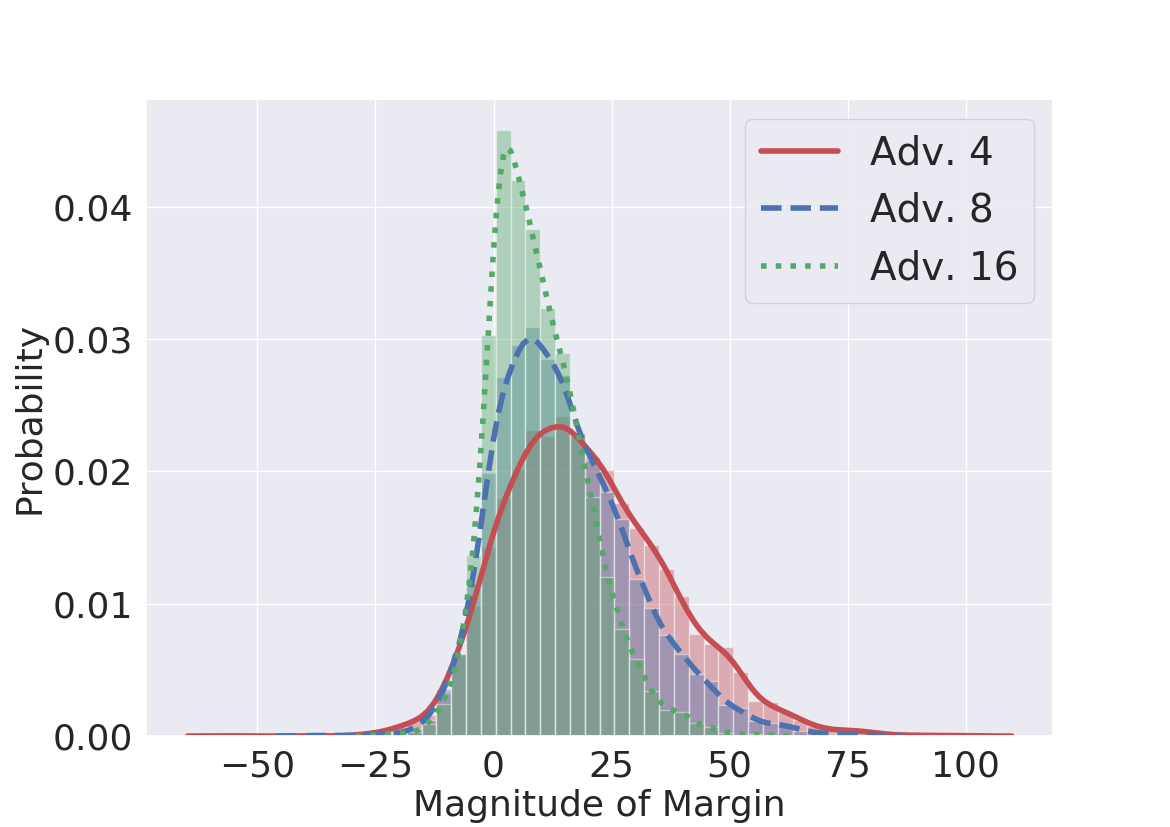}
      \caption{Margin Distribution}
      \label{fig:cifar10_maring_test_pre}
    \end{subfigure}
    \begin{subfigure}[b]{0.16\textwidth}
      \includegraphics[width=\columnwidth]{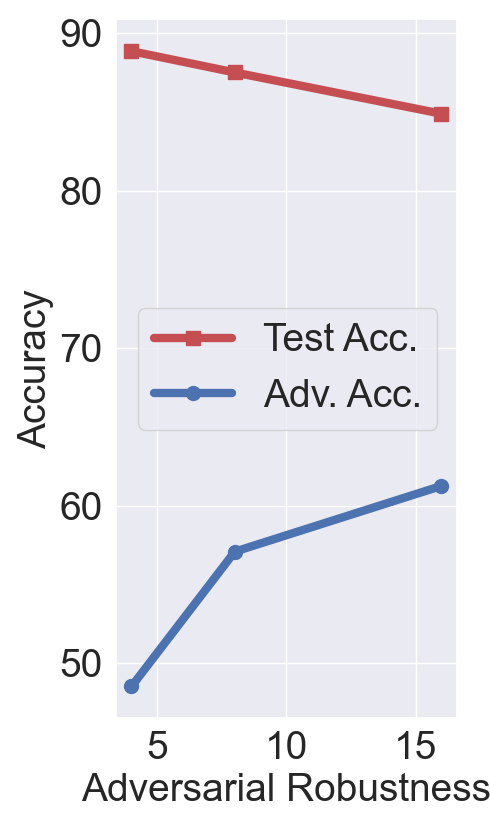}
      \caption{Accuracy}
      \label{fig:accuracy_pre}
    \end{subfigure}
    \caption[Adversarial Robustness Regularization Effect Demo]{
      Experiment results on ResNet56 \citep{He2016} trained on the CIFAR10 dataset.
      For the details of the experiments, refer to \cref{sec:exper-valid}.
      {\bf (a)} The standard deviation of singular values of each layer of NNs with adversarial robustness (AR) strength 4, 16 (AR strength 8 is dropped for clarity of the plot). 
      To emphasize, the $x$-axis is the layer index --- overall 56 layers are involved. 
      {\bf (b)} The probability distribution of margins of NNs with AR strength 4, 8, 16. {\bf (c)} The standard and adversarial accuracy of NNs with AR 4, 8, 16.
    }
    \label{fig:diffidence}
\end{figure*}

{\bf Key results.} 
Through a series of theoretically motivated experiments, we find that AR is achieved by regularizing/biasing NNs towards less confident solutions by making the changes in the feature space of most layers (which are induced by changes in the instance space) smoother uniformly in all directions; so to a certain extent, it prevents sudden change in prediction w.r.t. perturbations. 
However, the end result of such smoothing concentrates examples around decision boundaries and leads to worse standard performance. 
We elaborate the above statement in details shortly in
\cref{sec:advers-robusnt-leads}.

{\bf Implications.} We conjecture that the improper reduction comes from the
indistinguishability of the change induced in the intermediate layers of NNs by
adversarial noise and that by inter-class difference.  To guarantee AR, NNs are
asked to smoothe out difference uniformly in all directions in a high
dimensional space, and thus are biased towards less confident solutions that make
similar/concentrated predictions.  We leave the investigation of the conjecture
as future works.

\subsection{AR leads to less confident NNs with more indecisive misclassifications}
\label{sec:advers-robusnt-leads}

This section elaborates the {\it key results} we briefly present previously.

{\it AR reduces the perturbations in the activation/outputs --- the perturbations
that are induced by perturbations in the inputs fed into the layer --- of most layers.}
Through a series of theoretically motivated experiments, the results prompt us
to look at the singular value distributions of the weight matrix of each layer
of the NNs. Shown in \cref{fig:singular_value_CIFAR10_pre}, we find that
overall the standard deviation (STD) of singular values associated with a layer of the NN trained with lower AR strength 4 is larger than that of the NN with higher AR strength 16
\footnote{The AR strength is characterized by the maximally allowed $l_{\infty}$ norm of adversarial examples that are used to train the NNs --- we use adversarial training \citep{Madry2017} to build adversarial robustness into NNs. 
Details can be found in \cref{sec:techn-build-advers}} 
--- the green dots are mostly below the red dots.
Note that given a matrix $\bv{W}$ and an example $\bv{x}$, singular values of $\bv{W}$ determine how the norm $||\bv{Wx}||$ is changed comparing with $||\bv{x}||$. 
More specifically, let $\sigma_{\min}, \sigma_{\max}$ be the minimal and maximal singular values. 
If $\bv{x}$ is not in the null space of $\bv{W}$, then we have $||\bv{Wx}|| \in [\sigma_{\min}||\bv{x}||, \sigma_{\max}||\bv{x}||]$, where $||\cdot||$ denotes $2$-norm. This applies to norm $||\delta\bv{x} ||$ of a perturbation as well; that is, given possible changes $\delta{\bv{x}}$ of $\bv{x}$ of the same norm $||\delta\bv{x}|| = c$, where $c$ is a constant, the variance of $\sigma(\bv{W})$ roughly determines the variance of $||\bv{W}\delta\bv{x}||$, where $\sigma(\bv{W})$ denotes all singular values $\{\sigma_i\}$ of $\bv{W}$. 
In more details, note that by SVD decomposition, $\bv{W}\delta\bv{x} = \sum_{i}\sigma_{i}\bv{u}_i\bv{v}^T_i\delta\bv{x}$, thus $\sigma_i$ determines how the component $\bv{v}^{T}_i\delta\bv{x}$ in the direction of $\bv{v}_i$ is amplified. 
To see an example, suppose that $\sigma_{\min} = \sigma_{\max} = \sigma_0$, then the variance of $\sigma(\bv{W})$ is zero, and $||\bv{W}\delta\bv{x}|| = \sigma_0||\delta\bv{x}||$. 
In this case, the variance of $||\bv{W}\delta\bv{x}||$ (given an ensemble of perturbations $\delta\bv{x}$ of the same norm $c$) is zero as well. 
The conclusion holds as well for $\text{ReLU}(\bv{W}\delta{x})$, where $\bv{W}$ here
is a weight matrix of a layer of a NN, and $\text{ReLU}$ denotes Rectifier
Linear Unit activation function (proved by applying Cauchy interlacing law by
row deletion \citep{Meek} in \cref{lm:1}).  Consequently, by reducing the
variance of singular values of weight matrix of a layer of the NN, AR reduces
the variance of the norms of layer activations, or informally, perturbations in the
activations, induced by input perturbations.

{\it The perturbation reduction in activations concentrates examples, and it
empirically concentrates them around decision boundaries; that is, predictions
are less confident.}
The reduced variance implies that the outputs of each layer of the NN are more concentrated, but it does not tell where they are concentrated. 
Note that in the previous paragraph, the variance relationship discussed between $||\bv{W}\delta\bv{x}||$ and $||\delta\bv{x}||$ equally applies to $||\bv{W}\bv{x}||$ and $||\bv{x}||$, where $\bv{x}$ is an actual example instead of perturbations. 
Thus, to find out the concentration of perturbations, we can look at the
concentration of samples. 
Technically, we look at {\it margins} of examples. In a multi-class setting, suppose a NN computes a score function $f : \bb{R}^{d} \rightarrow \bb{R}^L$, where $L$ is the number of classes; a way to convert this to a classifier is to select the output coordinate with the largest magnitude, meaning $x \mapsto \argmax_{i}f_{i}(x)$. 
The {\it confidence} of such a classifier could be quantified by margins. It measures the gap between the output for the correct label and other labels, meaning $f_y(x) - \max_{i\not=y} f_i(x)$. 
Margin piece-wise linearly depends on the scores, thus the variance of margins is also in a piece-wise linear relationship with the variance of the scores, which are computed linearly from the activation of a NN layer. 
Thus, the consequence of concentration of activation discussed in the previous paragraph can be observed in the distribution of margins. 
More details of the connection between singular values and margins are discussed in \cref{sec:advers-robustn-makes}, after we present \cref{lm:1}.  
A zero margin implies that a classifier has equal propensity to classify an example to two classes, and the example is on the decision boundary.  
We plot the margin distribution of the test set of CIFAR10 in \cref{fig:cifar10_maring_test_pre}, and find that margins are increasingly concentrated around zero --- that is, the decision boundaries --- as AR strength grows.

{\it The sample concentration around decision boundaries smoothes sudden
changes induced perturbations, but also increases indecisive misclassifications.} 
The concentration of test set margins implies that the induced change in margins by the perturbation in the instance space is reduced by AR.
Given two examples $\bv{x}, \bv{x}'$ from the test set, $\delta\bv{x} = \bv{x} - \bv{x}'$ can be taken as a significant perturbation that changes the example $\bv{x}$ to $\bv{x}'$. 
The concentration of overall margins implies the change induced by $\delta\bv{x}$ is smaller statistically in NNs with higher AR strength. 
Thus, for an adversarial perturbation applied on $\bv{x}$, statistically the change of margins is smaller as well --- experimentally it is reflected in the increased adversarial robustness of the network, as shown in the increasing curve in \cref{fig:accuracy_pre}. 
That is, the sudden changes of margins originally induced by adversarial perturbations are {\it smoothed} (to change slowly). 
However, the {\it cost} of such smoothness is lower confidence in prediction, and more test examples are slightly/indecisively moved to the wrong sides of the decision boundaries --- incurring lower accuracy, as shown in the decreasing curve in \cref{fig:accuracy_pre}.
    
Lastly, we note that experiments in this section are used to illustrate our main arguments in this section.
Further consistent quality results are reported in \cref{sec:exper-valid} by
conducting experiments on CIFAR10/100 and Tiny-ImageNet with networks of varied
capacity. And more corroborative experiment results are presented in the
appendices, and outlined in \cref{sec:outl-contr}.

\subsection{Outline and contributions}
\label{sec:outl-contr}

This work carries out generalization analysis on NNs with AR.  
The quantities in the previous section are identified by the generalization errors (GE) upper bound we establish at \cref{thm:main}, which characterizes the regularization of AR on NNs. 
The key result is obtained at the {\it end} of a series of analysis, thus we present the outline of the analysis here.

{\bf Outline.} 
After presenting some preliminaries in \cref{sec:preliminaries}, we proceed to
analyze the regularization of AR on NNs, and establish a GE upper bound in
\cref{sec:gener-bound-guar}.  The bound prompts us to look at the GE gaps in
experiments.  In \cref{sec:regul-effects-nns}, we find that for NNs trained
with higher AR strength, the surrogate risk gaps (GE gaps) decrease for a range
of datasets, i.e., CIFAR10/100 and Tiny-ImageNet.  It implies AR effectively
regularizes NNs.  We then study the finer behavior change of NNs that might
lead to such a gap reduction.  Again, we follow the guidance of
\cref{thm:main}.  We look at the margins in \cref{sec:cms-that-concentrate},
then at the singular value distribution in \cref{sec:advers-robustn-makes}, and
discover the main results described in \cref{sec:advers-robusnt-leads}.  More
corroborative experiments are present in \cref{sec:further-evidence} and
\cref{sec:furth-empir-study} to show that such phenomenon exists in a broad
range of NNs with varied capacity and adversarial training techniques.  More
complementary results are present in \cref{sec:discr-trend-surr} to explain
some seemingly abnormal observations, and in \cref{sec:furth-evid-smooth} to
quantitatively demonstrate the smoothing effects of AR discussed in
\cref{sec:advers-robusnt-leads}.  Related works are present in
\cref{sec:related-works}.

{\bf Contributions}. 
Overall, the core contribution in this work is to show that adversarial robustness (AR) regularizes NNs in a way that hurts its capacity to learn to perform in test. 
More specifically:
\begin{itemize}
\item We establish a generalization error (GE) bound that characterizes the
regularization of AR on NNs. 
The bound connects \emph{margin} with adversarial robustness radius $\epsilon$ via \emph{singular values of weight matrices} of NNs, thus suggesting the two quantities that guide us to investigate the regularization effects of AR empirically.
\item Our empirical analysis tells that AR \emph{effectively} regularizes NNs to reduce the GE gaps. To understand how reduced GE gaps turns out to degrade test performance, we study \emph{variance of singular values} of layer-wise weight matrices of NNs and \emph{distributions of margins} of samples, when different strength of AR are applied on NNs.
\item  The study shows that AR is achieved by regularizing/biasing NNs
towards less confident solutions by making the changes in the feature space of
most layers (which are induced by changes in the instance space) smoother uniformly in all directions; so to a certain extent, it prevents sudden change in
prediction w.r.t. perturbations. 
However, the end result of such smoothing concentrates samples around decision boundaries and leads to worse standard performance.
\end{itemize}

%-------------------------------------------------------------------------
\section{Preliminaries}
\label{sec:preliminaries}

Assume an instance space $\mathcal{Z} = \mathcal{X} \times \mathcal{Y}$, where $\mathcal{X}$ is the space of input data, and $\mathcal{Y}$ is the label space.
$Z := (X, Y)$ are the random variables with an unknown distribution $\mu$, from which we draw samples. 
We use $S_m = \{ z_i = (\bvec{x}_i, y_i)\}_{i=1}^{m}$ to denote the training set of size $m$ whose examples are drawn independently and identically distributed (i.i.d.) by sampling $Z$. 
Given a loss function $l$, the goal of learning is to identify a function $T: \mathcal{X} \mapsto \mathcal{Y}$ in a hypothesis space (a class $\mathcal{T}$ of functions) that minimizes the expected risk
\[
  R(l \circ T) = \mathbb{E}_{Z \sim \mu}\left[l\left(T(X), Y\right)\right] ,
\]
Since $\mu$ is unknown, the observable quantity serving as the proxy to the expected risk $R$ is the empirical risk
\begin{displaymath}
  R_m(l \circ T) = \frac{1}{m}\sum\limits_{i=1}^{m}l\left(T(\bvec{x}_i), y_i\right).
\end{displaymath}
Our goal is to study the discrepancy between $R$ and $R_m$, which is termed as {\it generalization error} --- it is also sometimes termed as generalization gap in the literature
\begin{equation}
  \label{eq:ge}
  \text{GE}(l \circ T) = |R(l \circ T) - R_m(l \circ T)|.
\end{equation}

A NN is a map that takes an input $x$ from the space
$\mathcal{X}$, and builds its output by recursively applying a linear map
$W_{i}$ followed by a pointwise non-linearity $g$:
\[
  x_i = g (\bv{W}_{i} \bv{x}_{i-1}) ,
\]
where $i$ indexes the times of recursion, which is denoted as a layer in the
community, $i = 1, \ldots, L$, $x_0 = x$, and $g$ denotes the activation function.
which is restricted to Rectifier Linear Unit (ReLU) \citep{Glorot2011} or
max pooling operator \citep{Becigneul2017} in this paper. To compactly summarize the operation of $T$, we denote
\begin{equation}
  \label{eq:nn}
  Tx = g(\bv{W}_Lg(\bv{W}_{L-1}\ldots g(\bv{W}_1\bv{x}))) .
\end{equation}

\begin{definition}[Covering number]
    Given a metric space $(\mathcal{S}, \rho)$, and a subset $\tilde{\mathcal{S}} \subset \mathcal{S}$, we say that a subset $\hat{\mathcal{S}}$ of $\tilde{\mathcal{S}}$ is a $\epsilon$-cover of $\tilde{\mathcal{S}}$, if $\forall \tilde{s} \in \tilde{\mathcal{S}}$, $\exists \hat{s} \in \hat{\mathcal{S}}$ such that $\rho(\tilde{s}, \hat{s}) \leq \epsilon$. The $\epsilon$-covering number of $\tilde{\mathcal{S}}$ is
    \begin{displaymath}
        \mathcal{N}_{\epsilon}(\tilde{\mathcal{S}}, \rho) = \min\{ |\hat{\mathcal{S}}|:
        \hat{\mathcal{S}} \text{ is an } \epsilon\text{-covering of } \tilde{\mathcal{S}} \} .
    \end{displaymath}
\end{definition}

Various notions of adversarial robustness have been studied in existing works
\citep{Madry2017,Turner2018,Zhang2019}. 
They are conceptually similar; in this work, we formalize its definition to make clear the object for study.

\begin{definition}[$(\rho, \epsilon)$-adversarial robustness]
    \label{def:ar}
    Given a multi-class classifier $f: \ca{X} \rightarrow \bb{R}^{L}$, and a metric $\rho$ on $\ca{X}$, where $L$ is the number of classes, $f$ is said to be adversarially robust w.r.t. adversarial perturbation of strength $\epsilon$, if there exists an $\epsilon > 0$ such that $\forall z = (\bv{x}, y) \in \ca{Z}$ and $ \delta\bv{x} \in \{\rho(\delta \bv{x}) \leq \epsilon \}$, we have
    \begin{displaymath}
        f_{\hat{y}}(\bv{x} + \delta\bv{x}) - f_{i}(\bv{x} + \delta\bv{x}) \geq 0,
    \end{displaymath}
    where $\hat{y} = \argmax_{j}f_j(\bv{x})$ and $i \not= \hat{y} \in \ca{Y}$.
    $\epsilon$ is called \textit{\textbf{adversarial robustness radius}}.
    When the metric used is clear, we also refer $(\rho, \epsilon)$-adversarial robustness as $\epsilon$-adversarial robustness.
\end{definition}

Note that the definition is an {\it example-wise} one; that is, it requires each example to have a guarding area, in which all examples are of the same class. Also note that the robustness is w.r.t. the predicted class, since ground-truth label is unknown for a $\bv{x}$ in test.

We characterize the GE with ramp risk, which is a typical risk to undertake theoretical analysis \citep{Bartlett2017,Role2019}.

\begin{definition}[Margin Operator]
    \label{def:margin}
    A margin operator $\ca{M}: \bb{R}^{L} \times \{1, \ldots, L\} \rightarrow
    \bb{R}$ is defined as
    \begin{displaymath}
        \ca{M}(\bv{s}, y) := s_{y} - \max_{i\not= y}s_{i}
    \end{displaymath}
    \end{definition}
    \begin{definition}[Ramp Loss]
    \label{def:ramp_loss}
    The ramp loss $l_{\gamma}: \bb{R} \rightarrow \bb{R}^{+}$ is defined as
    \begin{displaymath}
        l_{\gamma} (r) :=
        \begin{cases}
        0            & r < -\gamma        \\
        1 + r/\gamma & r \in [-\gamma, 0] \\
        1            & r > 0
        \end{cases}
    \end{displaymath}
\end{definition}
\begin{definition}[Ramp Risk]
    \label{def:ramp_risk}
    Given a classifier $f$, ramp risk is the risk defined as
    \begin{displaymath}
        R_{\gamma}(f) := \bb{E}(l_{\gamma}(- \ca{M}(f(X), Y))),
    \end{displaymath}
    where $X, Y$ are random variables in the instance space $\ca{Z}$ previously.
\end{definition}

We will use a different notion of margin in \cref{thm:main}, and formalize its
definition as follows. We reserve the unqualified word ``margin'' specifically
for the margin discussed previously --- the output of margin operator for
classification. We call this margin to-be-introduced {\it instance-space margin
(IM)}.

\begin{definition}[Smallest Instance-space Margin]
    \label{def:im}
    Given an element $z = (\bv{x},
    y) \in \ca{Z}$, let $v(\bv{x})$ be the distance from $\bv{x}$ to its closest
    point on the decision boundary, i.e., the {\it instance-space
      margin} (IM) of example $\bv{x}$.
Given a covering set $\hat{S}$ of $\mathcal{Z}$, let
\begin{equation}
    \label{eq:smallest_margin}
  v_{\min} = \min_{\bv{x} \in \{\bv{x} \in \mathcal{X} |  \exists
    \bv{x}' \in \hat{S}_m, ||\bv{x} - \bv{x}'||_2 \leq \epsilon\}} v(\bv{x}),
\end{equation}
where $\hat{S}_m := \{\bv{x}' \in \hat{S} | \exists \bv{x}_i \in S_m, ||\bv{x}_i - \bv{x}'||_2 \leq \epsilon \}$.
    $v_{\min}$ is the {\it smallest} instance-space margin of elements in the
    covering balls that {\it contain training examples}.
\end{definition}

%-------------------------------------------------------------------------
\section{Theoretical instruments for empirical studies on AR}
\label{sec:gener-bound-guar}

In this section, we rigorously establish the bound mentioned in the introduction. 
We study the map $T$ defined in \cref{sec:preliminaries} as a NN (though technically, $T$ now is a map from $\ca{X}$ to $\bb{R}^{L}$, instead of to $\ca{Y}$, such an abuse of notation should be clear in the context). 
To begin with, we introduce an assumption, before we state the generalization error bound guaranteed by adversarial robustness.

\begin{assumption}[Monotony]
    \label{asp:monotony}
    Given a point $\bv{x} \in \ca{X}$, let $\bv{x}'$ be the point on the decision boundary of a NN $T$ that is closest to $\bv{x}$. 
    Then, for all $\bv{x}''$ on the line segment $\bv{x} + t(\bv{x'} - \bv{x}), t \in [0, 1]$, the margin $\ca{M}(T\bv{x}'', y)$ decreases monotonously.
\end{assumption}

The assumption is a regularity condition on the classifier that rules out undesired oscillation between $\bv{x}$ and $\bv{x}'$. 
To see how, notice that the margin defined in \cref{def:margin} reflects how confident the decision is made. 
Since $\bv{x}'$ is on the decision boundary, it means the classifier is unsure how it should be classified. 
Thus, when the difference $\bv{x}' - \bv{x}$ is gradually added to $\bv{x}$, ideally we want the confidence that we have on classifying $\bv{x}$ to decrease in a consistent way to reflect the uncertainty.

\begin{figure*}[t]
    \centering
    \begin{subfigure}[b]{0.48\textwidth}
      \def\svgwidth{\columnwidth}
      %% Creator: Inkscape inkscape 0.92.3, www.inkscape.org
%% PDF/EPS/PS + LaTeX output extension by Johan Engelen, 2010
%% Accompanies image file 'ar_demo.pdf' (pdf, eps, ps)
%%
%% To include the image in your LaTeX document, write
%%   \input{<filename>.pdf_tex}
%%  instead of
%%   \includegraphics{<filename>.pdf}
%% To scale the image, write
%%   \def\svgwidth{<desired width>}
%%   \input{<filename>.pdf_tex}
%%  instead of
%%   \includegraphics[width=<desired width>]{<filename>.pdf}
%%
%% Images with a different path to the parent latex file can
%% be accessed with the `import' package (which may need to be
%% installed) using
%%   \usepackage{import}
%% in the preamble, and then including the image with
%%   \import{<path to file>}{<filename>.pdf_tex}
%% Alternatively, one can specify
%%   \graphicspath{{<path to file>/}}
%% 
%% For more information, please see info/svg-inkscape on CTAN:
%%   http://tug.ctan.org/tex-archive/info/svg-inkscape
%%
\begingroup%
  \makeatletter%
  \providecommand\color[2][]{%
    \errmessage{(Inkscape) Color is used for the text in Inkscape, but the package 'color.sty' is not loaded}%
    \renewcommand\color[2][]{}%
  }%
  \providecommand\transparent[1]{%
    \errmessage{(Inkscape) Transparency is used (non-zero) for the text in Inkscape, but the package 'transparent.sty' is not loaded}%
    \renewcommand\transparent[1]{}%
  }%
  \providecommand\rotatebox[2]{#2}%
  \newcommand*\fsize{\dimexpr\f@size pt\relax}%
  \newcommand*\lineheight[1]{\fontsize{\fsize}{#1\fsize}\selectfont}%
  \ifx\svgwidth\undefined%
    \setlength{\unitlength}{578.88939565bp}%
    \ifx\svgscale\undefined%
      \relax%
    \else%
      \setlength{\unitlength}{\unitlength * \real{\svgscale}}%
    \fi%
  \else%
    \setlength{\unitlength}{\svgwidth}%
  \fi%
  \global\let\svgwidth\undefined%
  \global\let\svgscale\undefined%
  \makeatother%
  \begin{picture}(1,0.90206607)%
    \lineheight{1}%
    \setlength\tabcolsep{0pt}%
    \put(0.3343688,0.30556906){\color[rgb]{0,0,0}\makebox(0,0)[lt]{\lineheight{0}\smash{\begin{tabular}[t]{l} \end{tabular}}}}%
    \put(0,0){\includegraphics[width=\unitlength,page=1]{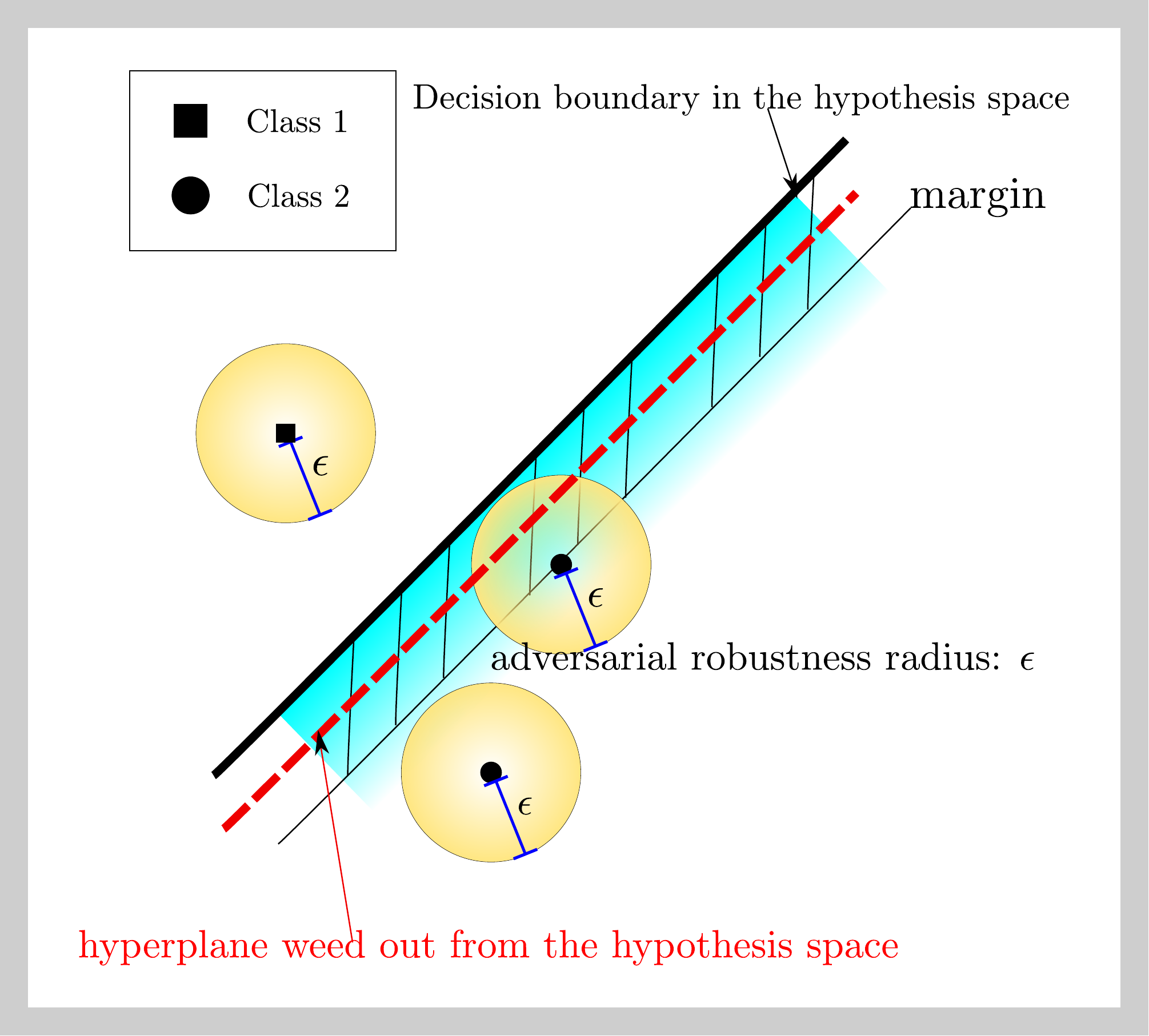}}%
  \end{picture}%
\endgroup%

      \caption{}
      \label{fig:demo}
    \end{subfigure}
    \begin{subfigure}[b]{0.44\textwidth}
      \def\svgwidth{\columnwidth}
      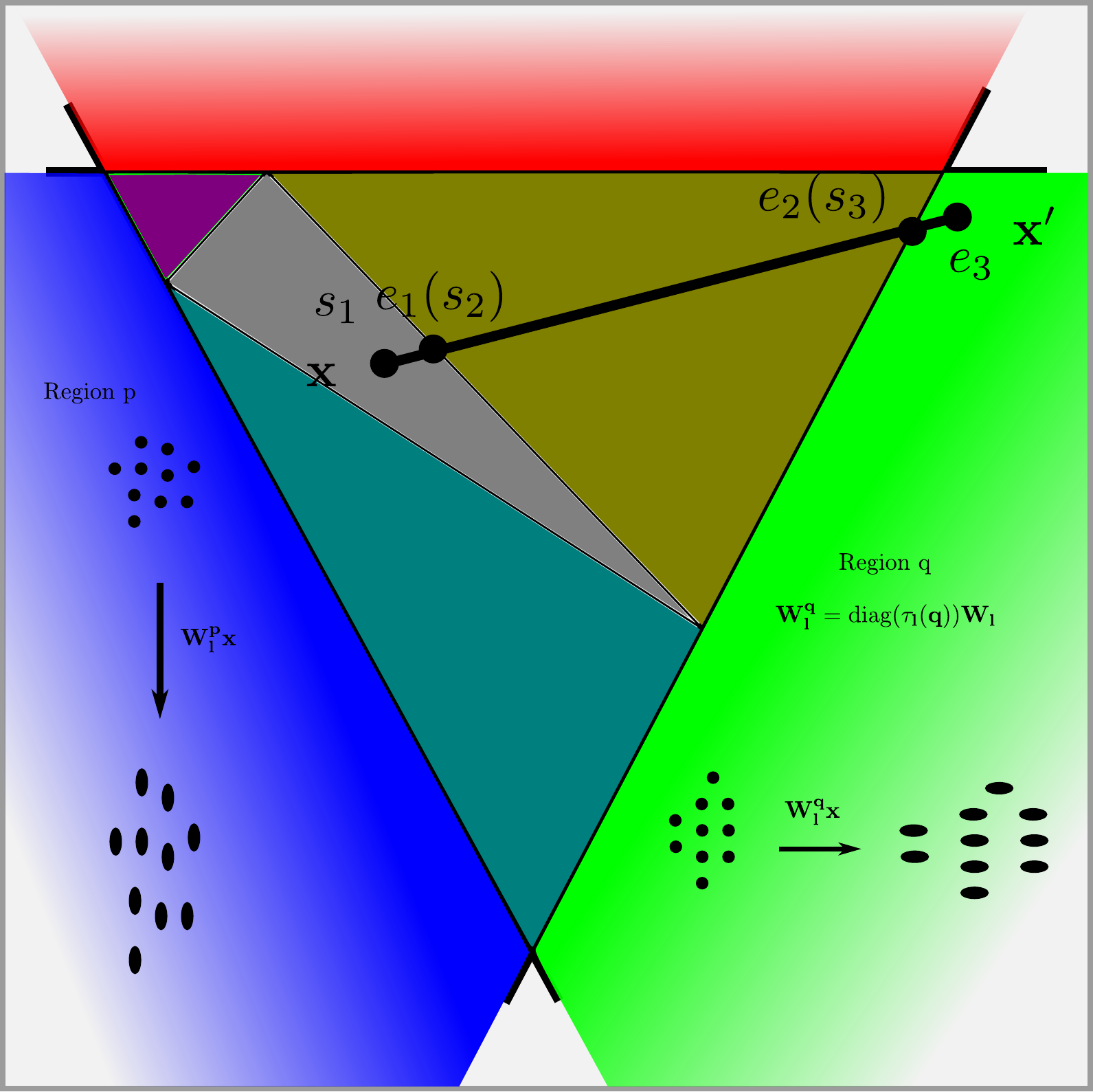
      \caption{}
      \label{fig:piecewise}
    \end{subfigure}
    \caption[
        Adversarial Robustness Regularization Effect Demo]{{\bf (a)} Illustration of the regularization effect of adversarial robustness. 
        If a NN $T$ is $\epsilon$-adversarially robust, for a given example $\bv{x}$ (drawn as filled squares or circles) and points $\bv{x}'$ in the yellow ball $\{\bv{x}' \,|\, \rho(\bv{x}, \bv{x}') \leq \epsilon\}$ around $\bv{x}$, the predicted labels of $\bv{x}, \bv{x}'$ should be the same, and the loss variation is potentially bigger as $\bv{x}'$ moves from the center to the edge, as shown as intenser yellow color at the edge of a ball. 
        Collectively, the adversarial robustness of each example requires an {\it instance-space margin} (IM) to exist for the decision boundary, shown as the shaded cyan margin. 
        As normally known, margin is related to generalization ability that shrinks the hypothesis space. 
        In this case, the IM required by adversarial robustness would weed out hypotheses that do not have an adequate IM, such as the red dashed line shown in the illustration. 
        {\bf (b)} Illustration of \cref{lm:1}. 
        Given a NN with ReLU activation function, the feature map $\bv{I}_l$ at layer $l$ is divided into regions where $\bv{I}_l(\bv{x})$ is piecewise linear w.r.t. $\bv{x}$. 
        The induced linear map $\bv{W}^q_1$ is given by $\text{diag}(\tau_1(q))\bv{W}_1$, where $\text{diag}(\tau_l(q))$ is a diagonal matrix whose diagonal entries are given by a vector $\tau_1(q)$ that has 0-1 values. 
        For example, in region $p$, $\bv{I}_1 = \bv{W}^{p}_1\bv{x}$ and distance between instances $\bv{x}$ are vertical elongated, while in region $q$, $\bv{I}_1 = \bv{W}^{q}_1\bv{x}$ and distance are horizontally elongated. 
        Thus given $\bv{x}, \bv{x}'$, the difference $\left|\left|  \bv{I}_{l}(\bv{x}) - \bv{I}_{l}(\bv{x}') \right|\right|$ between $\bv{I}_l(\bv{x})$ and $\bv{I}_l(\bv{x}')$ is the length of the transformed line segment $\bv{x} - \bv{x}'$ drawn, of which each segment is linearly transformed in a different way.
    }
\end{figure*}

\begin{theorem}
    \label{thm:main}
    Let $T$ denote a NN with ReLU and MaxPooling nonlinear activation functions (the definition is put at \cref{eq:nn} for readers' convenience), $l_{\gamma}$ the ramp loss defined at \cref{def:ramp_loss}, and $\ca{Z}$ the instance space assumed in \cref{sec:gener-bound-guar}. 
    Assume that $\ca{Z}$ is a $k$-dimensional regular manifold that accepts an $\epsilon$-covering with covering number $(\frac{C_{\ca{X}}}{\epsilon})^{k}$, and assumption \cref{asp:monotony} holds. 
    If $T$ is $\epsilon_0$-adversarially robust (defined at \cref{def:ar}), $\epsilon \leq \epsilon_0$, and denote $v_{\min}$ the smallest IM margin in the covering balls that contain training examples (defined at \cref{def:im}), $\sigma_{\min}^{i}$ the smallest singular values of weight matrices $\bv{W}_{i}, i=1,\ldots,L-1$ of a NN, $\{\bv{w}_{i}\}_{i=1,\ldots, |\ca{Y}|}$ the set of vectors made up with $i$th rows of $\bv{W}_{L}$ (the last layer's weight matrix), then given an i.i.d. training sample $S_m = \{ z_i = (\bvec{x}_i, y_i) \}_{i=1}^{m}$ drawn from $\ca{Z}$, its generalization error $\GE(l \circ T)$ (defined at \cref{eq:ge}) satisfies that, for any $\eta > 0$, with probability at least $1 - \eta$
    \begin{align}
        \GE(l_{\gamma} \circ T) \leq&  \max\{0, 1 - \frac{u_{\min}}{\gamma}\}\nonumber\\
                        &+
                        \sqrt{\frac{2\log(2)C_{\mathcal{X}}^{k}}{\epsilon^{k}m}
        + \frac{2 \log(1/\eta)}{m}}\label{eq:bound_1}
    \end{align}
    where
    \begin{equation}
        u_{\min} = \min_{y, \hat{y} \in \ca{Y}, y\not=\hat{y}}||\bv{w}_{y} - \bv{w}_{\hat{y}}||_2\prod_{i=1}^{L-1}\sigma_{\min}^{i}v_{\min}\label{eq:bound_2}
    \end{equation}
    is a lower bound of margins of examples in covering balls that
    contain training samples.
\end{theorem}
  
The proof of \cref{thm:main} is in \cref{sec:proof-}. {\it The bound identifies quantities that would be studied experimentally in \cref{sec:exper-valid} to understand the regularization of AR on NNs.}
The first term in \cref{eq:bound_1} in \cref{thm:main} suggests that quantities related to the lower bound of {\it margin} $u_{\min}$ might be useful to study how $\epsilon$-adversarial robustness ($\epsilon$-AR) regularizes NNs.
However, $\epsilon$-AR is guaranteed in the instance space that determines the smallest {\it instance-space margin} $v_{\min}$. 
To relate GE bound with $\epsilon$-AR, we characterize in \cref{eq:bound_2} the relationship between margin with IM, via smallest {\it singular values of NNs' weight matrices}, suggesting that quantities related to singular values of NNs' weight matrices might be useful to study how AR regularizes NNs as well.  
An illustration on how AR could influence generalization of NNs through IM is also given in \cref{fig:demo}.
The rightmost term in \cref{eq:bound_1} is a standard term in robust framework \citep{Xu2012a} in learning theory, and is not very relevant to the discussion.
{\it The remaining of this paper are empirical studies that are based on the quantities}, e.g., margin distributions and singular values of NNs' weight matrices, that are related to the identified quantities, i.e., $u_{\min}, \sigma_{\min}^{i}$. 
These studies aim to illuminate with empirical evidence on the phenomena that AR regularizes NNs, reduces GE gaps, but degrades test performance.
\footnote{ 
    Note that in the previous paragraph, though we identifies quantities $u_{\min}$ and $\sigma_{\min}^{i}$ related to the upper bound of GE, the quantities we actually would study empirically are {\it margin distribution} and all {\it singular values} that characterize the GE of all samples, not just the extreme case (upper bound). 
    The analytic characterization of the GE of all samples is not possible since we do not have enough information (we do not know the true distribution of samples). 
    That's why to arrive at close-form analytic characterization of GE, we resort to the extreme non-asymptotic large-sample behaviors. 
    {\it The analytic form is a neat way to present how relevant quantities influence GE.} 
    In the rest of the paper, we would carry on empirical study on the distributions of margins and singular values to investigate AR's influence on GE of all samples.
}

\begin{figure*}[t]
    \centering
    \begin{subfigure}[b]{0.49\textwidth}
      \includegraphics[width=\linewidth]{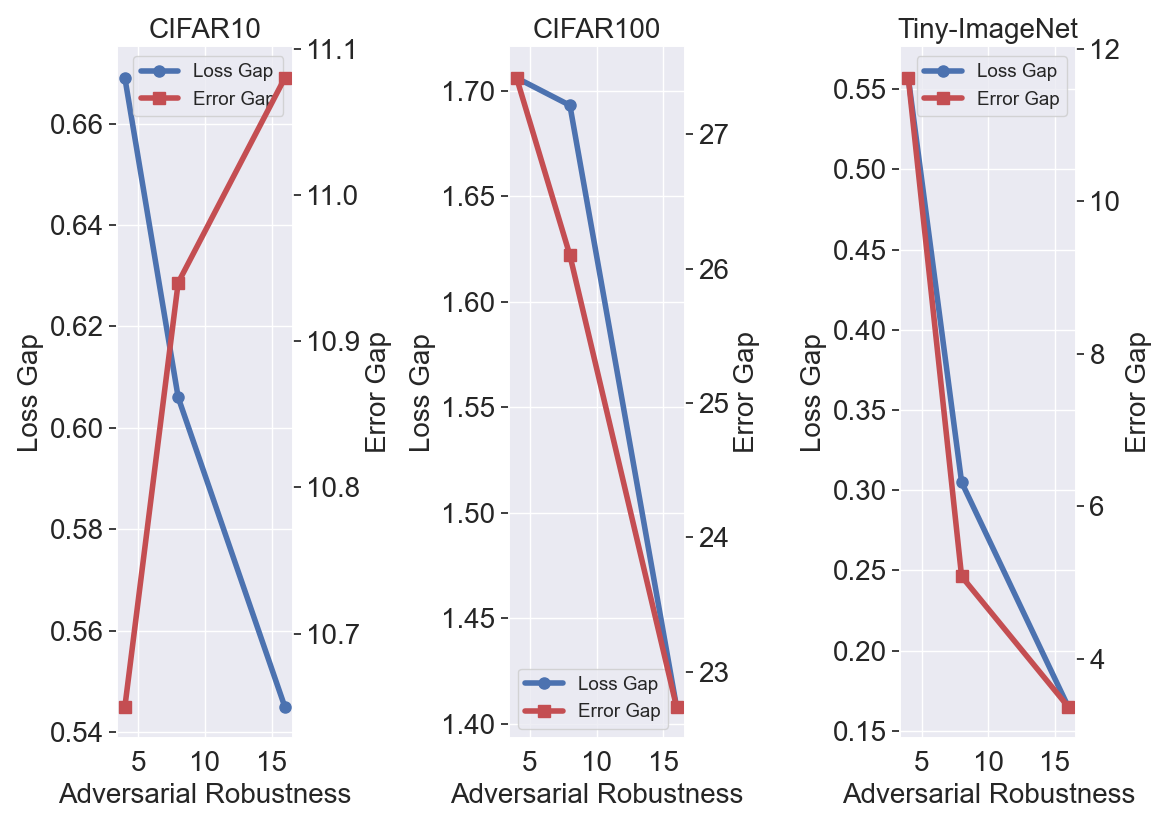}
      \caption{}
      \label{fig:loss_gap_full}
    \end{subfigure}
    \begin{subfigure}[b]{0.49\textwidth}
      \includegraphics[width=\linewidth]{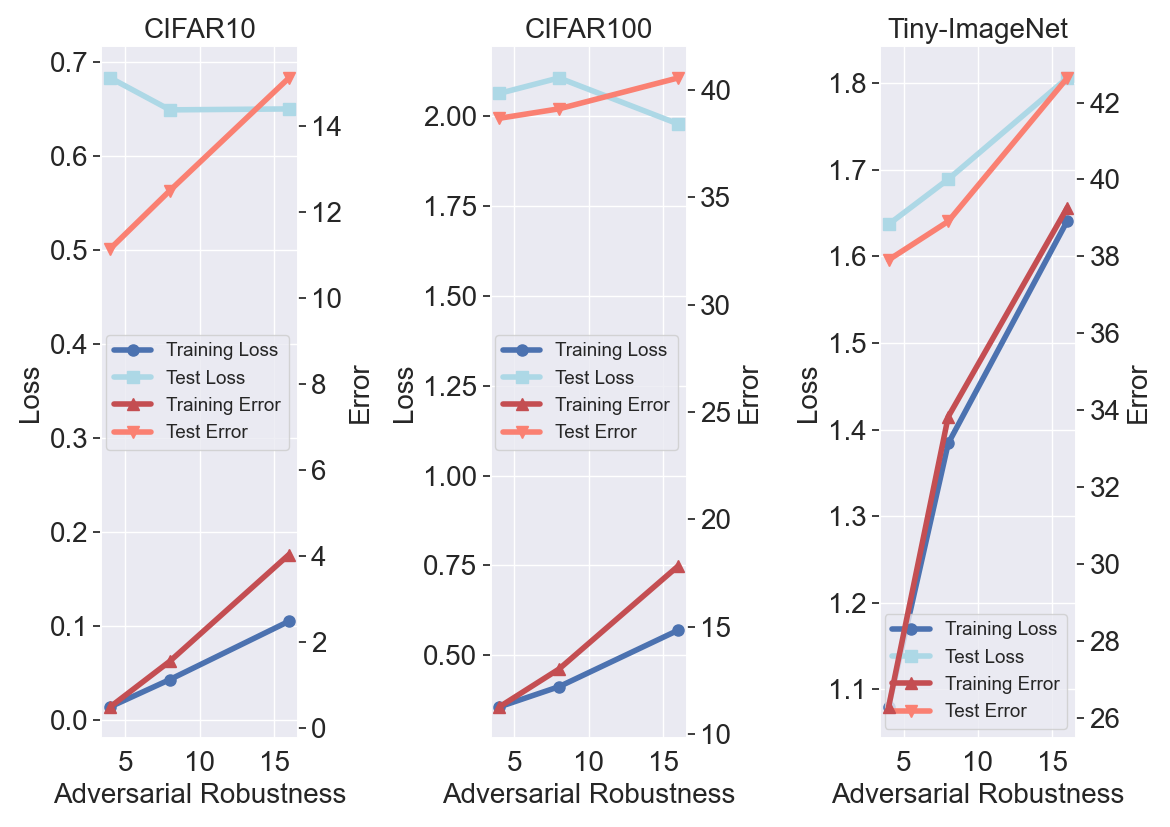}
      \caption{}
      \label{fig:performance}
    \end{subfigure}
    \caption{
        Experiment results on CIFAR10/100, and Tiny-ImageNet.  The unit of
x-axis is the adversarial robustness (AR) strength of NNs, c.f. the beginning
of \cref{sec:exper-valid}.  {\bf (a)} Plots of loss gap (and error rate gap)
between training and test datasets v.s. AR strength. {\bf (b)} Plots of losses
(and error rates) on training and test datasets v.s. AR strength.
        }
    \label{fig:datasets}
\end{figure*}

Before turning into empirical study, we further present a lemma to illustrate the relation characterized in \cref{eq:bound_2} without the need to jump into proof of \cref{thm:main}. 
It would motivate our experiments later in \cref{sec:advers-robustn-makes}. 
We state the following lemma that relates distances between elements in the instance space with those in the feature space of any intermediate network layers.
\begin{lemma}
\label{lm:1}
    Given two instances $\bv{x}, \bv{x}' \in \ca{X}$, let $\bv{I}_{l}(\bv{x})$ be the activation $g(\bv{W}_lg(\bv{W}_{l-1}\ldots g(\bv{W}_1\bv{x})))$ at layer $l$ of $\bv{x}$, then there exist $n \in \bb{N}$ sets of matrices $\{\bv{W}_i^{q_{j}}\}_{i=1\ldots l}, j=1\ldots n$, that each of the matrix $\bv{W}_{i}^{q_j}$ is obtained by setting some rows of $\bv{W}_{i}$ to zero, and $\{q_j\}_{j=1\ldots n}$ are
    arbitrary distinctive symbols indexed by $j$ that index $\bv{W}^{q_j}_i$,  such that
    \begin{displaymath}
        \left|\left|  \bv{I}_{l}(\bv{x}) - \bv{I}_{l}(\bv{x}') \right|\right| =
        \sum_{j=1}^{n}\int_{s_j}^{e_j}\left|\left|\prod_{i=1}^{l}\bv{W}_{i}^{q_j}dt(\bv{x} - \bv{x}')\right|\right|
    \end{displaymath}
    where $s_{1}=0, s_{j+1}=e_{j}, e_{n}=1, s_j,e_j \in [0, 1]$ --- each $[s_j, e_j]$ is a segment in the line segment parameterized by $t$ that connects $\bv{x}$ and $\bv{x}'$.
\end{lemma}
Its proof is in \cref{sec:proof-}, and an illustration is given in \cref{fig:piecewise}. 
Essentially, it states that difference in the feature space of a NN, induced by the difference between elements in the instance space, is a summation of the norms of the linear transformation ($\prod_{i=1}^{l}\bv{W}^{q_j}_i$) applied on segments of the line segment that connects $\bv{x}, \bv{x}'$ in the instance space. 
Since $\bv{W}_{i}^{q_j}$ is obtained by setting rows of $\bv{W}_i$ to zero, the singular values of these induced matrices are intimately related to weight matrices $\bv{W}_i$ of NN by Cauchy interlacing law by row deletion \citep{Meek}. 
Since the margin of an example $\bv{x}$ is a linear transform of the difference between $I_{L-1}(\bv{x})$ and the $I_{L-1}(\bv{x}')$ of an element $\bv{x}'$ on the decision boundary, singular values of $\{W_{i}\}_{i=1\ldots L-1}$ determine the amplification/shrinkage of the IM $\bv{x} - \bv{x}'$.

\section{Empirical studies on regularization of adversarial robustness}
\label{sec:exper-valid}

In this section, guided by \cref{thm:main}, we undertake empirical studies to explore AR's regularization effects on NNs. We first investigate the behaviors of off-the-shelf architectures of fixed capacity on various datasets in \cref{sec:regul-effects-nns} and \cref{sec:refin-anals-regul}. 
More corroborative controlled studies that explore the regularization effects of AR on NNs with varied capacity are present in \cref{sec:regul-effects-nns-1}.

\subsection{Adversarial robustness effectively regularizes NNs on various datasets}
\label{sec:regul-effects-nns}

This section aims to explore whether AR can effectively reduce generalization
errors --- more specifically, the surrogate risk gaps. 
We use adversarial training \citep{Madry2017} to build adversarial robustness into NNs.  
The AR strength is characterized by the maximally allowed $l_{\infty}$ norm of adversarial examples that are used to train the NNs. 
Details on the technique to build adversarial robustness into NNs is given in \cref{sec:techn-build-advers}.

Our experiments are conducted on CIFAR10, CIFAR100, and Tiny-ImageNet \citep{tiny_imagenet} that represent learning tasks of increased difficulties.
We use ResNet-56 and ResNet-110 \citep{He2016} for CIFAR10/100, and Wide ResNet (WRN-50-2-bottleneck) \citep{Zagoruyko2016} for Tiny-ImageNet \citep{tiny_imagenet}. 
These networks are trained with increasing AR strength. 
Results are plotted in \cref{fig:datasets}.

\paragraph{Regularization of AR on NNs.}
We observe in \cref{fig:loss_gap_full} (shown as blue lines marked by circles) that GE gaps (the gaps between training and test losses) decrease as strength of AR increase; we also observe in \cref{fig:loss_gap_full} that training losses increase as AR strength increase; these results (and more results in subsequent \cref{fig:control}) imply that AR does regularize training of NNs by reducing their capacities to fit training samples. 
Interestingly, in the CIFAR10/100 results in \cref{fig:performance}, the test losses show a decreasing trend even when test error rates increase. 
It suggests that the network actually performs better measured in test loss as contrast to the performance measured in test error rates. 
This phenomenon results from that less confident wrong predictions are made by NNs thanks to adversarial training, which will be explained in details in \cref{sec:refin-anals-regul}, when we carry on finer analysis. 
We note that on Tiny-ImageNet, the test loss does not decrease as those on CIFAR10/100. 
It is likely because the task is considerably harder, and regularization hurts NNs even measured in test loss.

\paragraph{Trade-off between regularization of AR and test error rates.}
The error rate curves in \cref{fig:performance} also tell that the end result of AR regularization leads to biased-performing NNs that achieve degraded test performance. 
These results are consistent across datasets and networks.

\paragraph{Seemingly abnormal phenomenon.}
An seemingly abnormal phenomenon in CIFAR10 observed in \cref{fig:loss_gap_full} is that the error rate gap actually increases. 
It results from the same underlying behaviors of NNs, which we would introduce in \cref{sec:refin-anals-regul}, and an overfitting phenomenon that AR cannot control. Since it would be a digress to explain, it is put in \cref{sec:discr-trend-surr}.

We finally note that the adversarial robustness training reproduced is relevant, of which the defense effect is comparable with existing works.
One may refer to \cref{fig:defense} in \cref{sec:exp_1} for the details. 
We can see from it that similar adversarial robustness to \citet{Madry2017} and \citet{Li2018a} is achieved for CIFAR10/100, Tiny-ImageNet in the NNs we reproduce.

\subsection{Refined analysis through margins and singular values}
\label{sec:refin-anals-regul}

The experiments in the previous sections confirm that AR reduces GE, but decreases accuracy. 
We study the underlying behaviors of NNs to analyze what have led to it here. 
More specifically, we show that adversarial training implements $\epsilon$-adversarial robustness by making NNs biased towards less confident solutions; that is, the key finding we present in \cref{sec:advers-robusnt-leads} that explains both the prevented sudden change in prediction w.r.t. sample perturbation (i.e., the achieved AR), and the reduced test accuracy.

\subsubsection{Margins that concentrate more around zero lead to reduced GE gap}
\label{sec:cms-that-concentrate}

To study how GE gaps are reduced, \cref{thm:main} suggests we first look at the
margins of examples --- a lower bound of margins is $u_{\min}$ in
\cref{eq:bound_2}.  The analysis on margins has been a widely used tool in
learning theory \citep{Bartlett2017}. It reflects the confidence that a
classifier has on an example, which after being transformed by a loss function,
is the surrogate loss.  Thus, the loss difference between examples are
intuitively reflected in the difference in confidence characterized by margins.
To study how AR influences generalization of NNs, distributions of samples
which are obtained by training ResNet-56 on CIFAR10 and CIFAR100 with increased
AR strength (the same setting as for \cref{fig:datasets}).  Applying the same
network of ResNet-56 respectively on \emph{CIFAR-10 and CIFAR-100 of different
learning difficulties} creates learning settings of \emph{larger- and
smaller-capacity} NNs.

\begin{figure}[h]
    \centering
    \begin{subfigure}[b]{0.235\textwidth}
      \includegraphics[width=\columnwidth]{./Fig/Probability_Distribution_of_Classification_Margin_on_CIFAR10_Test_Set_vanillaP_test.png}
      \caption{CIFAR10 Test}
      \label{fig:cifar10_margin_test}
    \end{subfigure}
    \begin{subfigure}[b]{0.235\textwidth}
      \includegraphics[width=\columnwidth]{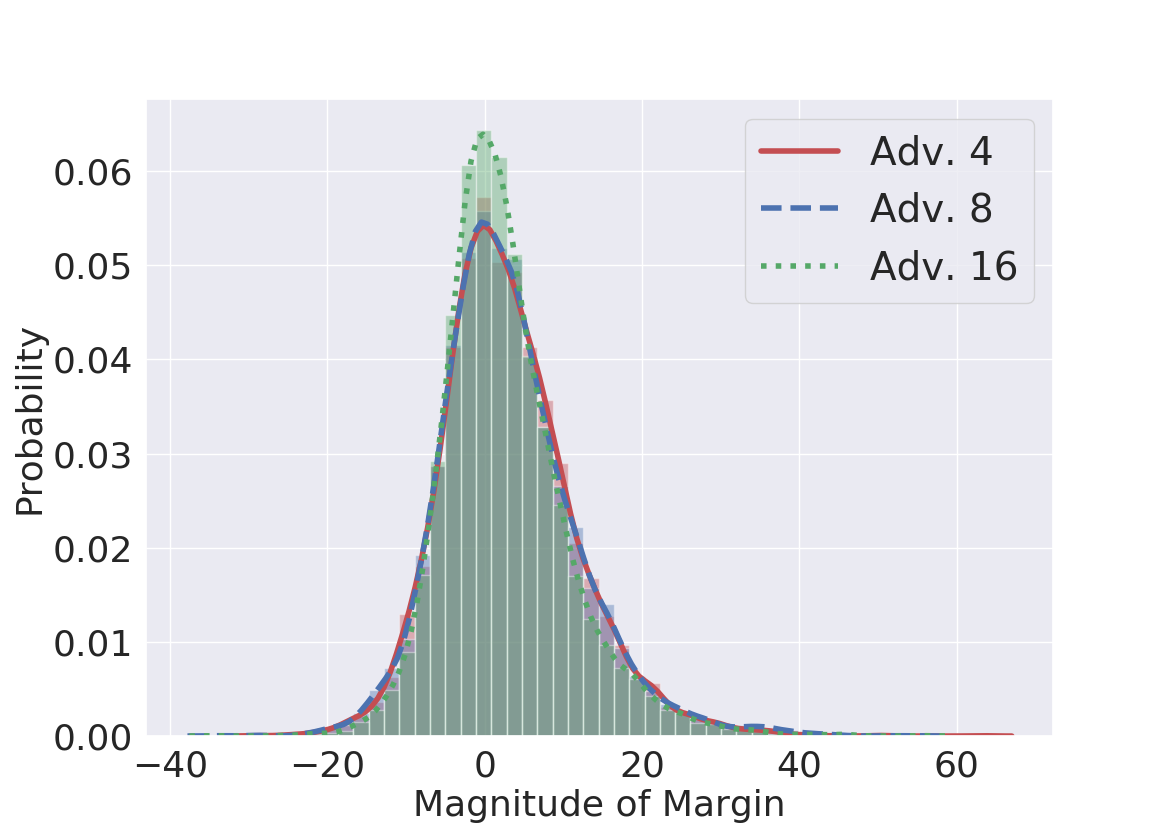}
      \caption{CIFAR100 Test}
      \label{fig:cifar100_margin_test}
    \end{subfigure}
    \\
    \begin{subfigure}[b]{0.235\textwidth}
      \includegraphics[width=\columnwidth]{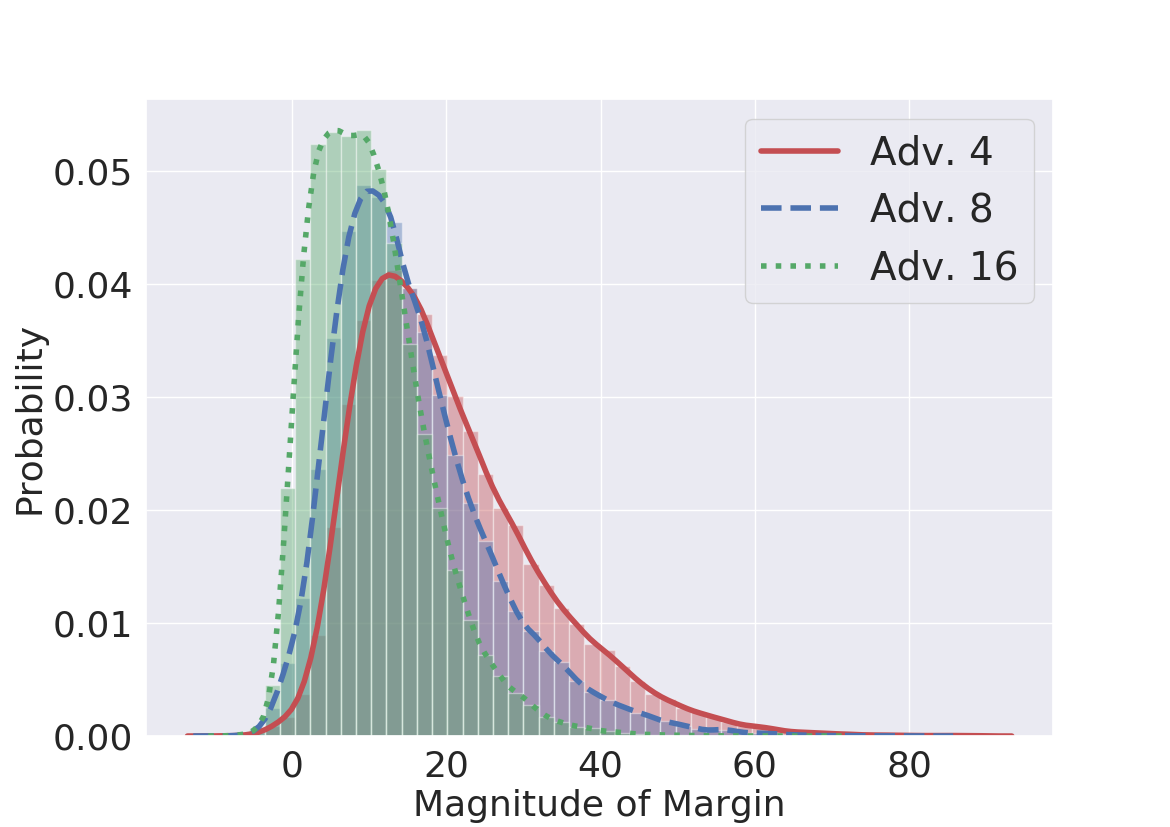}
      \caption{CIFAR10 Training}
      \label{fig:cifar10_margin_train}
    \end{subfigure}
    \begin{subfigure}[b]{0.235\textwidth}
      \includegraphics[width=\columnwidth]{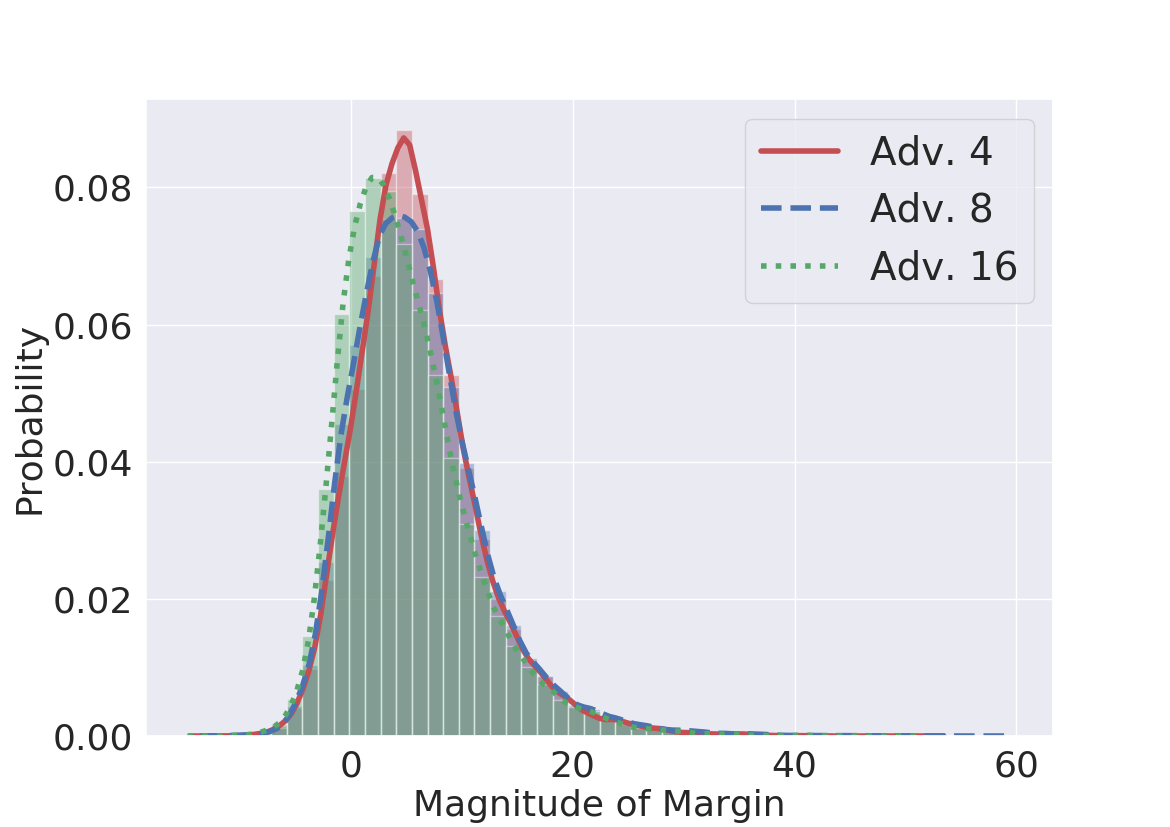}
      \caption{CIFAR100 Training}
      \label{fig:cifar100_margin_train}
    \end{subfigure}
    \caption{Margin distributions of NNs with AR strength 4, 8, 16 on Training and
      Test sets of CIFAR10/100.}
      \label{fig:margin_distribution}
\end{figure}

\paragraph{Concentration and reduced accuracy.}
In \cref{fig:margin_distribution}, we can see that in both CIFAR10/100, the
distributions of margins become more concentrated around zero as AR grows. The
concentration moves the mode of margin distribution towards zero and more
examples slightly across the decision boundaries, where the margins are zero,
which explains the reduced accuracy \footnote{We remark a possibly confusing
  phenomenon here about the margin. The bound \cref{eq:bound_1} might give the
  impression that a smaller margin might lead to a larger generalization error,
  while the empirical study instead shows that the NNs with a smaller margin have
  a smaller generalization error. The hypothesized confusion is a
  misunderstanding of the generalization bound analysis.
The upper bound is a worst case analysis of GE. However, in practice, the interesting object is the average gap between
the training losses and the test losses, i.e., the GE. Unfortunately, the
average gap cannot be analyzed analytically (cf. footnote 2). Thus, we, and also the statistical
learning community, resort to worst case analysis to find an upper bound on GE
to identify quantities that might influence GE. In this case, the phenomenon suggests that the bound might be loose, though this is a
problem that plagues the statistical learning community \cite{Nagarajan2019}. But our focus
in this work is not to derive tight bounds, or reach definite conclusions from
bounds alone, but to guide experiments with the bound.}.

\paragraph{Concentration and reduced loss/GE gap.}
The concentration has different consequences on training and test
losses. Before describing the consequences, to directly relate the concentration
to loss gap, we further introduce {\it estimated probabilities} of
examples. This is because though we use ramp loss in theoretical analysis, in
the experiments, we explore the behaviors of more practically used {\it cross
entropy loss}. The loss maps one-to-one to estimated probability, but not to
margin, though they both serve as a measure of confidence.  Suppose
$\bv{p}(\bv{x})$ is the output of the {\it softmax} function of dimension $L$
($L$ is the number of target classes), and $y$ is the target label. The
estimated probability of $\bv{x}$ would be the $y$-th dimension of
$(\bv{p}(\bv{x}))$, i.e., $(\bv{p}(\bv{x}))_y$.
{\bf On the training sets}, since the NNs are optimized to perform well on the
sets, only a tiny fraction of them are classified wrongly. 
To concentrate the margin distribution more around zero, is to make almost all
of predictions that are correct less confident. 
Thus, a higher expected training loss ensues.
{\bf On the test sets}, the estimated probabilities of the target class
concentrate more around middle values, resulting from lower confidence/margins in
predictions made by NNs, as shown in \cref{fig:prob_histogram} (but the
majority of values are still at the ends).  Note that wrong predictions away
from decision boundaries (with large negative margins) map to
large loss values in the surrogate loss function. Thus, though NNs with larger
AR strength have lower accuracy, they give more predictions whose estimated
probabilities are at the middle (compared with NNs with smaller AR
strength). These predictions, even if relatively more of them are wrong, maps
to smaller loss values, as shown in \cref{fig:loss_histogram}, where we plot
the histogram of loss values of test samples. In the end, it results in
expected test losses that are lower, or increase in a lower rate than the
training losses on CIFAR10/100, Tiny-ImageNet, as shown in
\cref{fig:performance}. {\bf The reduced GE gap} results from
the increased training losses, and decreased or less increased test losses.

\subsubsection{AR makes NNs smoothe predictions w.r.t. input
  perturabtions in all directions}
\label{sec:advers-robustn-makes}

\begin{figure}[h]
  \centering
  \begin{subfigure}[b]{0.235\textwidth}
    \includegraphics[width=\columnwidth]{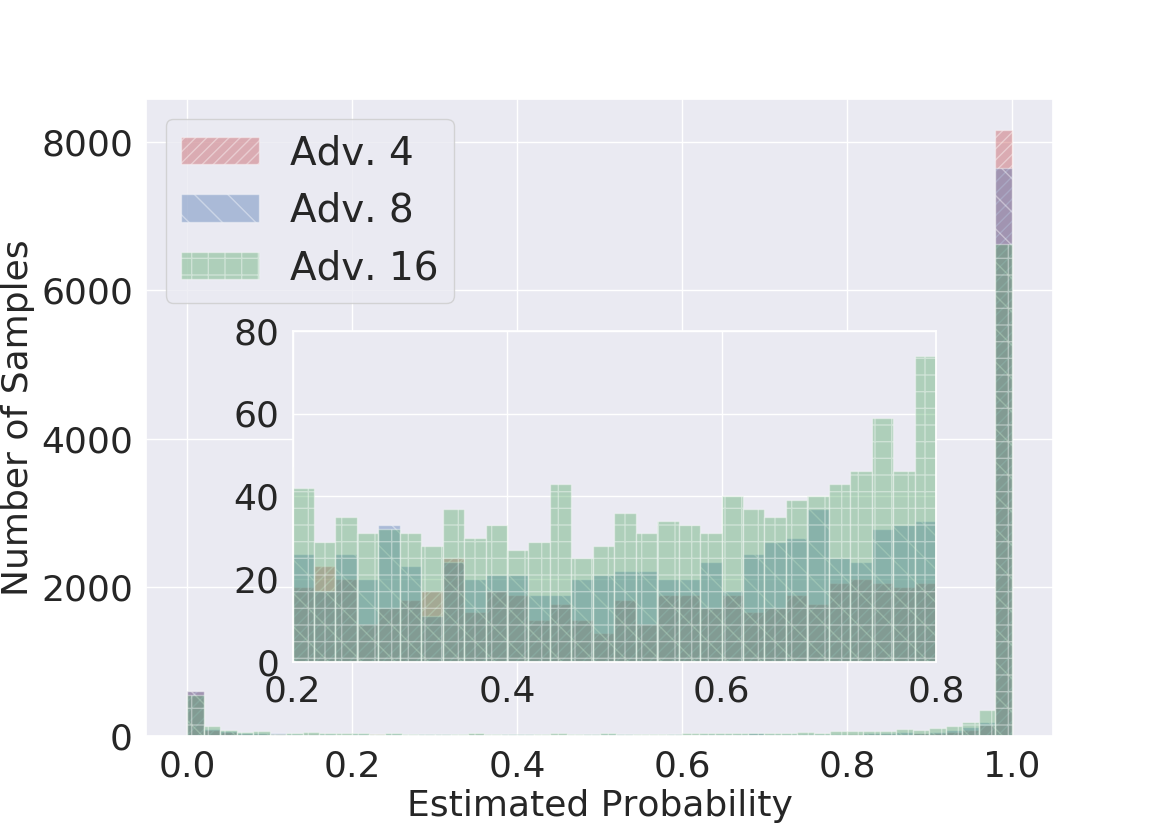}
    \caption{Prob. Histogram}
    \label{fig:prob_histogram}
  \end{subfigure}
  \begin{subfigure}[b]{0.235\textwidth}
    \includegraphics[width=\columnwidth]{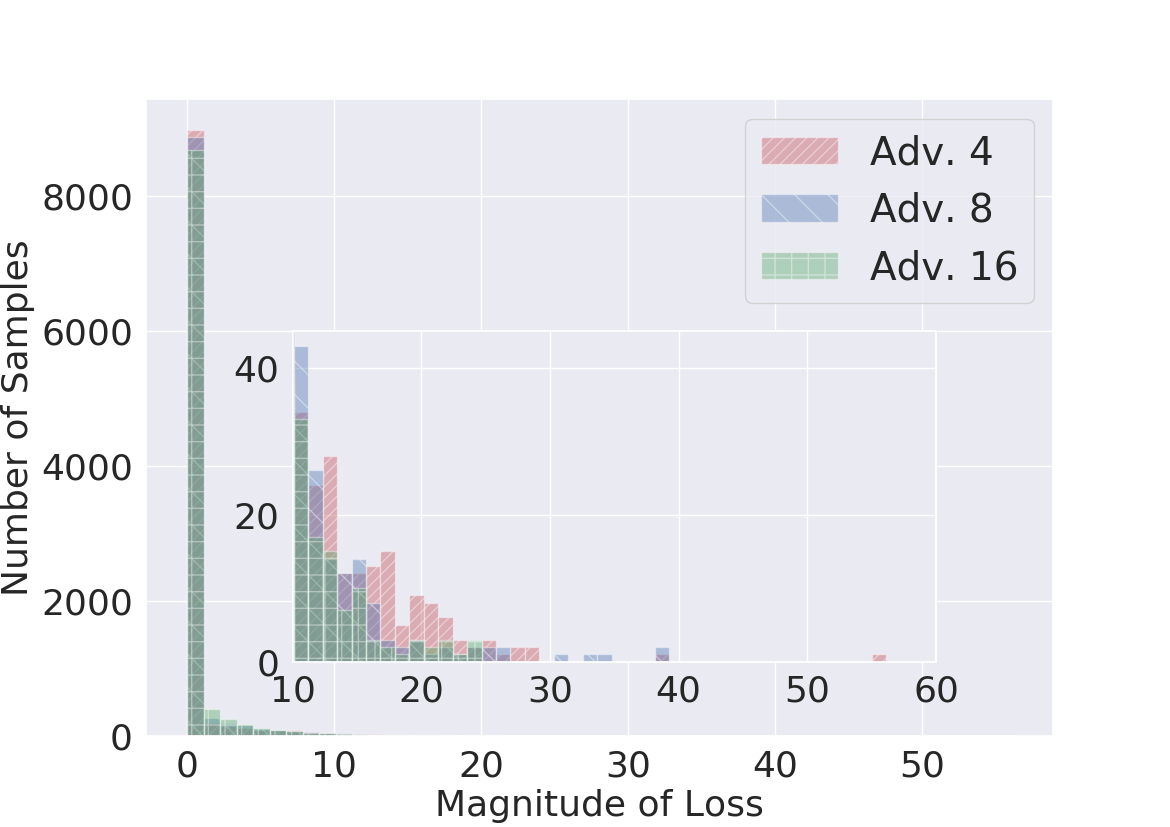}
    \caption{Loss Histogram}
    \label{fig:loss_histogram}
  \end{subfigure}
  \\
  \begin{subfigure}[b]{0.235\textwidth}
    \includegraphics[width=\columnwidth]{./Fig/Singular_value_var_on_CIFAR10.png}
    \caption{CIFAR10}
    \label{fig:singular_value_CIFAR10}
  \end{subfigure}
  \begin{subfigure}[b]{0.235\textwidth}
    \includegraphics[width=\columnwidth]{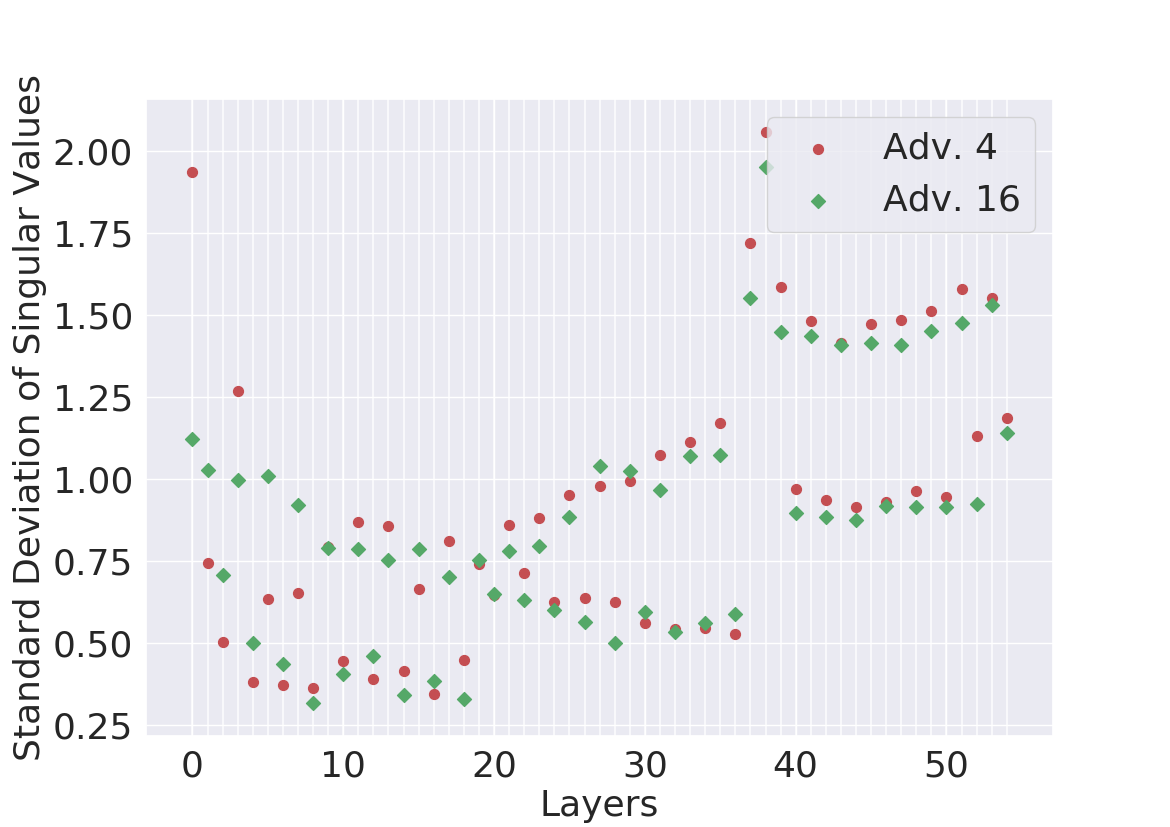}
    \caption{CIFAR100}
    \label{fig:singular_value_CIFAR100}
  \end{subfigure}
  \caption{{
    \bf (a)(b)} are histograms of estimated probabilities and losses of the test set sample of NNs trained with AR strength 4, 8, 16. 
    We plot a subplot of a narrower range inside the plot of the full range to better show the histograms of examples that are around the middle values induced by AR. 
    {\bf (c)(d)} are standard deviations of singular values of weight matrices of NNs at each layer trained on CIFAR10/100 with AR strength 4, 16. The AR strength 8 is dropped for clarity.}
    \label{fig:misc}
\end{figure}

The observation in \cref{sec:cms-that-concentrate} shows that AR make NNs just {\it less confident} by reducing the variance of predictions made and concentrate margins more around zero. 
In this section, we study the {\it underlying factors} of AR that make NNs become less confident.

To begin with, we show that the singular values of the weight matrix of each layer determine the perturbation in margins of samples induced by perturbations in the instance space.  
Such a connection between singular values and the perturbation of outputs of a single layer, i.e., $\text{ReLU}(\bv{W}\delta\bv{x})$, has been discussed in \cref{sec:advers-robusnt-leads}.
In the following, with \cref{lm:1}, we describe how the relatively more complex connection between margins and singular values of each weight matrix of layers of NNs holds. 
Observe that margins are obtained by applying a piece-wise linear mapping (c.f. the margin operator in \cref{def:margin}) to the activation of the last layer of a NN. 
It implies the perturbations in activation of the last layer induce changes in margins in a piece-wise linear way. 
Meanwhile, the perturbation in the activation of the last layer (induced by perturbation in the instance space) is determined by the weight matrix's singular values of each layer of NNs. 
More specifically, this is explained as follows. 
\Cref{lm:1} shows that the perturbation $\delta\bv{I}$ induced by $\delta\bv{x}$, is given by $\sum_{j=1}^{n}\int_{s_j}^{e_j}\left|\left|\prod_{i=1}^{l}\bv{W}_{i}^{q_j}\delta\bv{x}dt\right|\right|$.
Note that for each $i$, $\bv{W}^{q_i}_{i}$ is a matrix. By Cauchy interlacing law by row deletion \citep{Meek}, the singular values of $\bv{W}_{i}$, the weight matrix of layer $i$, determine the singular values of $\bv{W}^{q_j}_i$. 
Thus, suppose $l=1$, we have the change (measured in norm) induced by perturbation as $\sum_{j=1}^{n}\int_{s_j}^{e_j}\left|\left|\bv{W}_{1}^{q_j}\delta\bv{x}dt\right|\right|$. 
The singular values of $\bv{W}_{1}$ would determine the variance (of norms) of
activation perturbations induced by perturbations $\delta\bv{x}$, similarly as explained in
\cref{sec:advers-robusnt-leads} except that the norm perturbation now is obtained by
a summation of $n$ terms $\left|\left|\bv{W}_{1}^{q_j}\delta\bv{x}dt\right|\right|$
(each of which is the exact form discussed in \cref{sec:advers-robusnt-leads})
weighted by $1/(e_j - s_j)$.  Similarly, for the case where $l=2\ldots L-1$, the
singular values of $\bv{W}_{l}$ determine the variance of perturbations in the
output of layer $l$ that induced by the perturbations in the output of the
previous layer (the input to layer $l$), i.e., layer $l-1$.  Consequently, we
choose to study these singular values.

We show the standard deviation of singular values of each layer of ResNet56 trained on CIFAR10/100 earlier in \cref{fig:singular_value_CIFAR10} \cref{fig:singular_value_CIFAR100}.  
Overall, the standard deviation of singular values associated with a layer of the NN trained with AR strength 4 is mostly larger than that of the NN with AR strength 16.  
The STD reduction in CIFAR100 is relatively smaller than CIFAR10, since as observed in \cref{fig:cifar100_margin_test}, the AR induced concentration effect of margin distributions is also relatively less obvious than that in \cref{fig:cifar10_margin_test}. 
More quantitative analysis is given in \cref{sec:quant-analys-vari}.
This leads us to our {\it key results} described in \cref{sec:advers-robusnt-leads}.

\section*{Acknowledgements}

This work is supported in part by National Natural Science Foundation of China
(Grant No. 61771201), Program for Guangdong Introducing Innovative and
Enterpreneurial Teams (Grant No. 2017ZT07X183), Guangdong R\&D Key Project of
China (Grant No. 2019B010155001) and Guangzhou Key Laboratory of Body Data
Science (Grant No. 201605030011).

%-------------------------------------------------------------------------
%-------------------------------------------------------------------------

\bibliography{library,books,url,extra}
\bibliographystyle{icml2020}

\clearpage
\appendix
\section*{Appendices}
\label{sec:appendices}

%\pagebreak
%-------------------------------------------------------------------------

%\pagebreak
%-------------------------------------------------------------------------
\section{Related works}
\label{sec:related-works}\label{sec:furth-relat-works}

{\bf{Generalization and robustness.}}
Robustness in machine learning models is a large field. 
We review some more works that analyze robustness from the statistical perspective.
The majority of works that study adversarial robustness from the generalization
perspective study the generalization behaviors of machine learning models under
{\it adversarial risk}. 
The works that study adversarial risk include \citet{Attias2015,Schmidt,Cullina2018,Yin2018,Khim2018,Sinha2017}. 
The bounds obtained under the setting of adversarial risk characterize the risk gap introduced by adversarial examples.
Thus, it is intuitive that a larger risk gap would be obtained for a larger allowed perturbation limit $\epsilon$, which is roughly among the conclusions obtained in those bounds. 
That is to say, the conclusion normally leads to a larger generalization error as an algorithm is asked to handle more adversarial examples, for that it focuses on characterizing the error of adversarial examples, not that of natural examples. 
However, adversarial risk is not our focus. 
In this paper, we study when a classifier needs to accommodate adversarial examples, what is the influence that the accommodation has on generalization behaviors of empirical risk of natural data.

{\bf{Hard and soft adversarial robust regularization.}}
We study the behaviors of NNs that are trained in the way that adversarial
examples are required to be classified correctly. We note that the adversarial
robustness required can also be built in NNs in a soft way by adding a penalty
term in the risk function. Relevant works includes
\citet{Lyu2016} and \citet{Miyato2018a}. This line of works is not our subject of
investigation. They focus on increasing test performance instead of defense
performance. The focus of our works is to study the behaviors that lead to
standard performance degradation when a network is trained to has a reasonable
defense ability to adversarial examples. For example, a 50\% accuracy on adversarial
examples generated by PGD methods \citep{Madry2017} in \cref{fig:defense} is a
defense ability that can serve as a baseline for a reasonable defense
performance. It is natural that in the setting where the requirement to defend
against adversarial examples is dropped, the regularization can be weakened
(added as a penalty term) to {\it only} aim to improve the test performance of
the network. In this case, no performance degradation would occur, but the
defense performance is also poor.

{\bf Explicit regularization that increases robustness of NNs by imposing
smoothness through a penalty term.} The smoothing effect of adversarial
training on the loss surface has been observed in contemporary works
\cite{Moosavi-Dezfooli2019,Qin2019}. And based on such an observation, explicit
regularization is formulated by adding a penalty term to the risk function to
increase NNs' robustness.  The message of this work is different from the
insights of \cite{Moosavi-Dezfooli2019,Qin2019} related to regularization. They
\cite{Moosavi-Dezfooli2019,Qin2019} show that if the output of NNs is
explicitly smoothed through a penalty term thorough curvature regularization
\cite{Moosavi-Dezfooli2019}, or local linearization \cite{Qin2019}, then a
certain degree of adversarial robustness (AR) can be achieved. The penalty term
works as a regularizer because it is explicitly formulated that way. It is not
clear whether adversarial training, which is a different and arguably the most
widely used technique, has the effect of a regularizer. This is the issue that
is investigated in this work, and we show that adversarial training effectively
regularizes NNs, which is not clear previously. In addition, this work has
shown that adversarial training has a smoothing effect on features of all
layers, instead of just the loss surface. Such a fine-grained analysis is
possible because of the theoretical instruments developed in this work, and is
absent previously.

%\pagebreak
%-------------------------------------------------------------------------
\section{Further empirical studies on adversarial robustness}
\label{sec:furth-empir-stud}

\subsection{Technique to build adversarial robustness}
\label{sec:techn-build-advers}

To begin with, we describe the technique that we use to build AR into NNs.  
As mentioned in the caption of \cref{fig:diffidence}, we choose arguably the most well received technique, i.e., the adversarial training method \citep{Madry2017}.  
Specifically, we use $l_{\infty}$-PGD \citep{Madry2017} untargeted attack adversary, which creates an adversarial example by performing projected gradient descent starting from a random perturbation around a natural example. 
Then, NNs are trained with adversarial examples. 
NNs with different AR strength are obtained by training them with increasingly stronger adversarial examples. 
The adversarial strength of
adversarial examples is measured in the $l_{\infty}$ norm of the perturbation
applied to examples.  $l_{\infty}$-norm is rescaled to the range of $0-255$ to
present perturbations applied to different datasets in a comparable way; that
means in \cref{fig:diffidence} \cref{fig:datasets} \cref{fig:margin_distribution} \cref{fig:misc} and \cref{fig:control} \cref{fig:defense}, AR is measured in this scale. 
We use 10 steps of size 2/255 and maximum of = [4/255, 8/255, 16/255]
respectively for different defensive strength in experiments. For example, a NN
with AR strength $8$ is a NN trained with adversarial examples generated by
perturbations whose $l_{\infty}$ norm are at most $8$. Lastly, we note that although
adversarial training could not precisely guarantee an adversarial robustness
radius of $\epsilon$, a larger $l_{\infty}$ norm used in training would make NNs adversarially
robust in a larger ball around examples. Thus, though the precise adversarial
robustness radius is not known, we know that we are making NNs adversarially
robust w.r.t. a larger $\epsilon$. Consequently, it enables us to study the influence
of $\epsilon$-AR on NNs by studying NNs trained with increasing $l_{\infty}$ norm.

\subsection{Quantitative analysis of variance reduction in singular values}
\label{sec:quant-analys-vari}

Here, we provide more quantitative analysis on \cref{fig:singular_value_CIFAR10} 
and \cref{fig:singular_value_CIFAR100}, as noted previously in \cref{sec:advers-robustn-makes}.

Quantitatively, we can look at the accumulated standard
deviation (STD) difference in all layers. We separate the layers into two
group: the group that the STD (denoted $\sigma^4_i$) of singular values of layer $i$ (of
the NN trained) with AR strength 4 that is larger than that (denoted
$\sigma^{16}_{i}$) of AR strength 16; and the group that is smaller. In CIFAR10,
for the first group, the summation of the difference/increments of STD of the
two networks ($\sum_{i}\sigma^4_i - \sigma^{16}_i$) is 4.7465, and the average is 0.1158. For
the second groups, the summation ($\sum_{i}\sigma^{16}_i - \sigma^{4}_i$) is 0.4372, and
the average is 0.0312. In CIFAR100, the summation of the first group is 3.7511, and
the average is 0.09618; the summation of the second group is 0.4372, and the
average is 0.1103. The quantitative comparison shows that the accumulated STD
decrease in layers that have their singular value STDs decreased (comparing STD
of the NN with AR strength 16 with STD of the NN with AR strength 4) is {\it a
magnitude larger} the accumulated STD increase in the layers
that have their singular value STDs increased. The magnitude difference is
significant since the STDs of singular values of most layers are around $1$.

\subsection{Discrepancy between trends of loss and error rate gaps
in large capacity NNs}
\label{sec:discr-trend-surr}
In \cref{sec:regul-effects-nns}, we have noted an inconsistent behaviors of
CIFAR10, compared with that of CIFAR100 and Tiny-ImageNet: the error gap
reduces for CIFAR100 and Tiny-ImageNet, but increases for CIFAR10. It might
suggest that AR does not effectively regularize NNs in the case of
CIFAR10. However, we show in this section that the abnormal behaviors of CIFAR10 are
derived from the same margin concentration phenomenon observed in \cref{sec:cms-that-concentrate} due to capacity difference, and compared with
the error gaps, the GE/loss gaps are more faithfully representatives of the
generalization ability of the NNs. Thus, the seemingly abnormal phenomenon
corroborate, not contradict, the \emph{key results} present in \cref{sec:introduction}.

Using CIFAR10 and CIFAR100 as examples and evidence in the previous sections,
we explain how the discrepancy emerges from AR's influence on margin
distributions of the same network trained on tasks with different difficulties.
Further evidence that the discrepancy arises from capacity difference would be
shown at \cref{sec:regul-effects-nns-1}, where we run experiments to
investigate GE gap of NNs with varied capacities on the same task/dataset.

\label{sec:regul-effects-nns-1}

\begin{figure*}[t]
  \centering
  \begin{subfigure}[b]{0.49\textwidth}
    \includegraphics[width=\columnwidth]{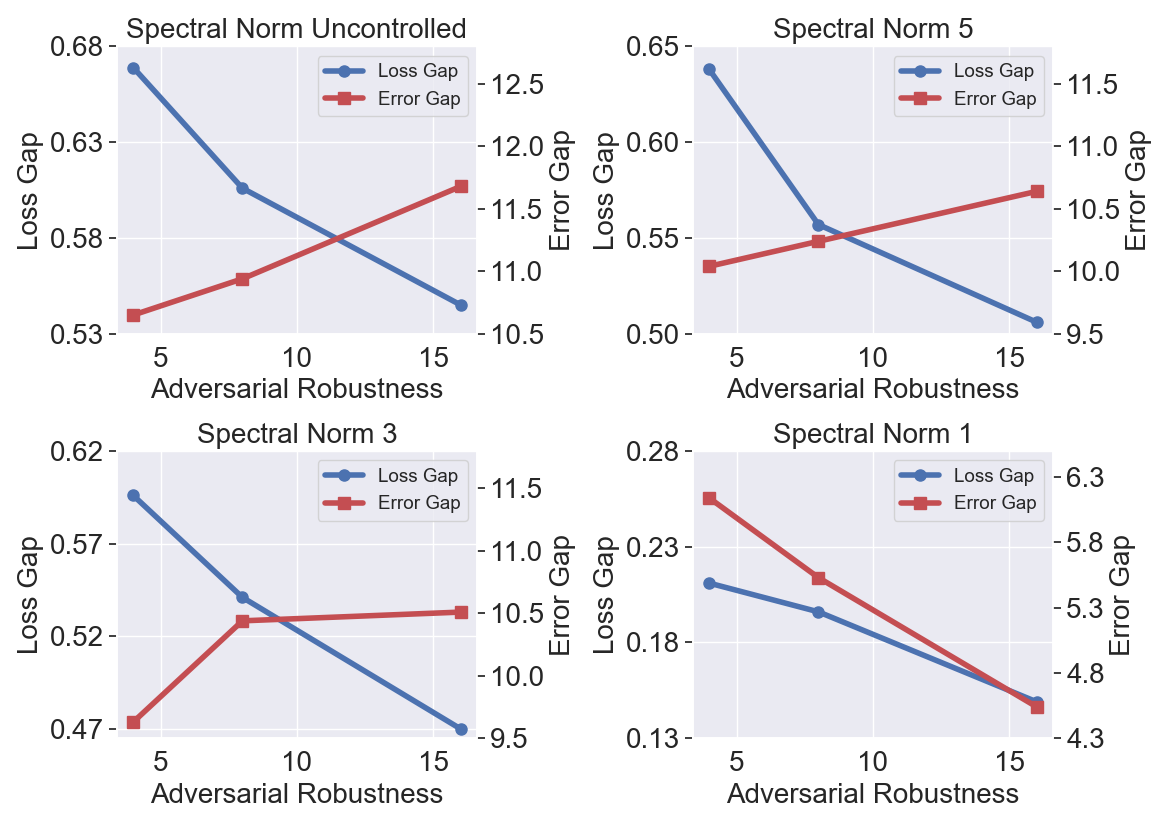}
    \caption{Gap Curves}
    \label{fig:control_resnet56}
  \end{subfigure}
  \begin{subfigure}[b]{0.49\textwidth}
    \includegraphics[width=\columnwidth]{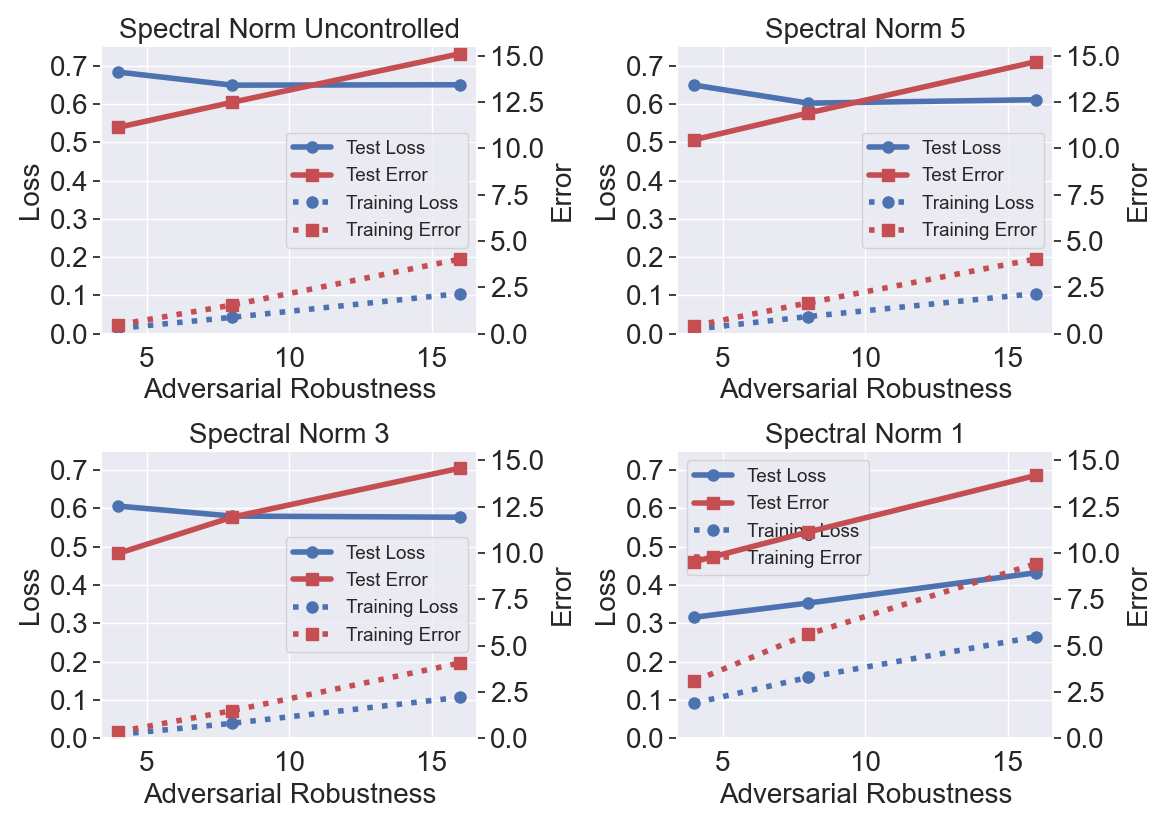}
    \caption{Error Rate \& Loss Curves}
    \label{fig:control_performance_resnet56}
  \end{subfigure}
  \caption{The four plots from upper left to lower bottom (in each subfigure)
are NNs with increasingly smaller spectral complexity, where ``Spectral Norm
1'' means for each weight matrix of the NN, its spectral norm is at most 1.
{\bf (a)} Plots of training/test loss gap (and error gap) against adversarial
robustness strength.  {\bf (b)} Training/test losses and error rates against
increased strength of adversarial robustness.
}
    \label{fig:control}
\end{figure*}

\begin{enumerate}
    \item
    {\it On CIFAR10, the margin distribution of training sets not only concentrate more around zero, but also skews towards zero.}
    As shown in the margin distribution on training sets of CIFAR10 in \cref{fig:cifar10_margin_train}, we find that the large error gap is caused by the high training accuracy that is achieved with a high concentration of training samples just slightly beyond the decision boundary. 
    This phenomenon does not happen in CIFAR100. Comparing margin distribution on the test set in fig. 4(a) in \cref{fig:cifar10_margin_test}, the margin distribution on the training set in \cref{fig:cifar10_margin_train} is highly skewed, i.e., asymmetrically distributed w.r.t. mean. While the margin distributions of CIFAR100 training set in \cref{fig:cifar100_margin_train} is clearly less skewed, and looks much more like a normal distribution, as that of the margin distribution on the test set.
    \item
    {\it The high skewness results from the fact that the NN trained on CIFAR10 is of
    large enough capacity to overfit the training set.}  As known, CIFAR100 is a
    more difficult task w.r.t. CIFAR10 with more classes and less training examples
    in each class. Thus, relatively, even the same ResNet56 network is used, the
    capacity of the network trained on CIFAR10 is larger than the one trained on
    CIFAR100. Recall that NNs have a remarkable ability to overfit training samples
    \citep{Zhang2016b}. And note that though AR requires in a ball around an
    example, the examples in the ball should be of the same class, since the ball
    is supposed only to include imperceptible perturbation to the example, few of
    the training samples are likely in the same ball. Thus, the ability to overfit
    the training set is not regularized by AR: if NNs can overfit all training
    samples, it can still overfit some more examples that are almost imperceptibly
    different.  For CIFAR10, since NNs have enough capacity, the NN simply overfits
    the training set.
    \item
    However, as shown in the observed overfitting phenomenon in \cref{fig:cifar10_margin_train}, the high training accuracy is made up of correct predictions with relatively lower confidence (compared with NNs with lower AR), which is bad and not characterized by the error rate; and the low test accuracy are made up of wrong predictions with relatively lower confidence as well (as explain in \cref{sec:cms-that-concentrate}), which is good, and not characterized by error rate as well. {\it Thus, the error gap in this case does not characterize the generalization ability (measured in term of prediction confidence) of NNs well, while the GE gap more faithfully characterizes the generalization ability, and show that AR effectively regularizes NNs.} 
    In the end, AR still leads to biased poorly performing solutions --- since the overfitting in training set does not prevent the test margin distribution concentrating more around zero, which leads to higher test errors of CIFAR10 as shown in \cref{fig:performance}.
    It further suggests that the damage AR done to the hypothesis space is not recovered by increasing capacity, however the ability of NNs to fit arbitrary labels is not hampered by AR.
\end{enumerate}

\subsection{Further evidence of regularization effects on NNs with varied capacity}
\label{sec:further-evidence}

In previous sections, we observe AR consistently effectively regularizes NNs;
meanwhile, we also observe that in the case where a NN has a large capacity, it
can spuriously overfit training samples and lead to an increased error gap.  In
this section, we present additional results by applying AR to networks of
varied capacities. The effects of adversarial training on a larger NNs, i.e.,
ResNet 110 is given in \cref{sec:regul-effects-nns-2}. Then, AR applied on NNs with controlled capacities
through spectral normalization is given in \cref{sec:regul-effects-nns-3}.
This is to ensure that our observations and analysis in previous sections exist
not just at some singular points, but also in a continuous area in the
hypothesis space.

\subsubsection{Regularization effects on NNs with larger capacity}
\label{sec:regul-effects-nns-2}

\begin{figure*}[t]
    \centering
    \begin{subfigure}[b]{0.49\textwidth}
      \includegraphics[width=\linewidth]{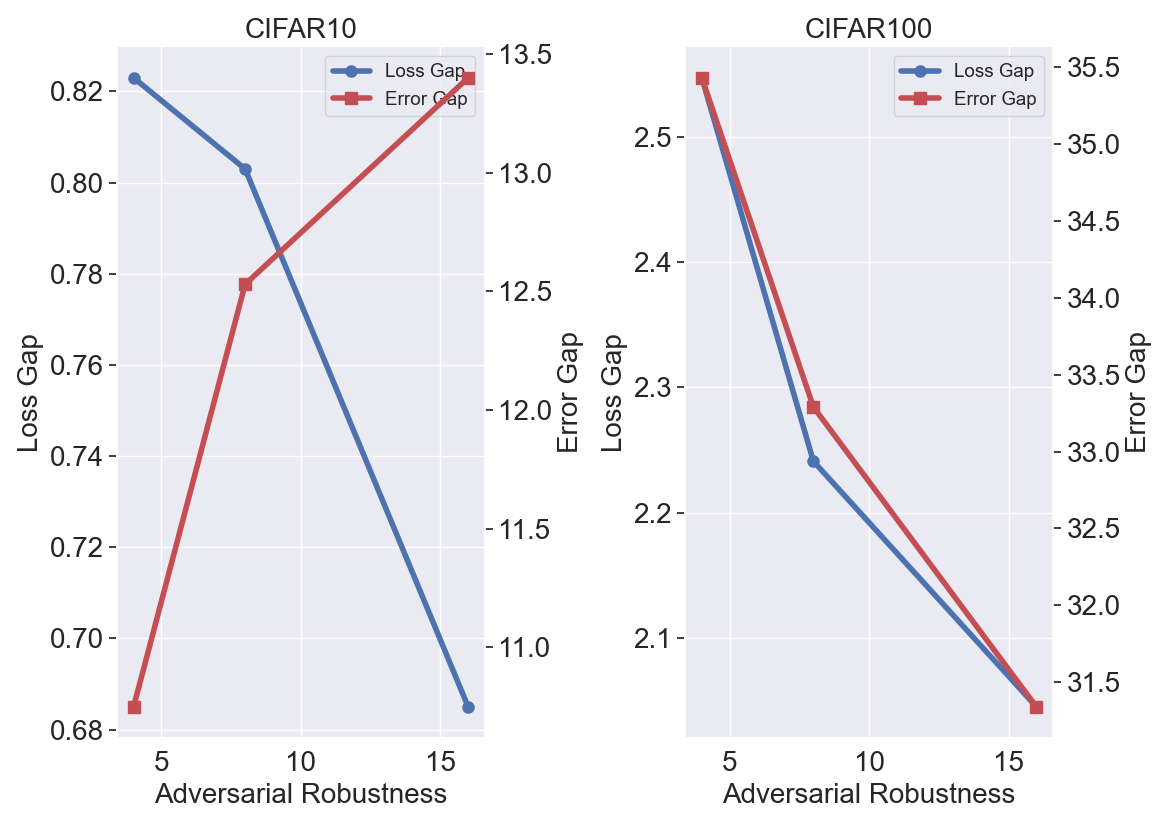}
      \caption{}
      \label{fig:loss_gap_full_resnet110}
    \end{subfigure}
    \begin{subfigure}[b]{0.49\textwidth}
      \includegraphics[width=\linewidth]{ 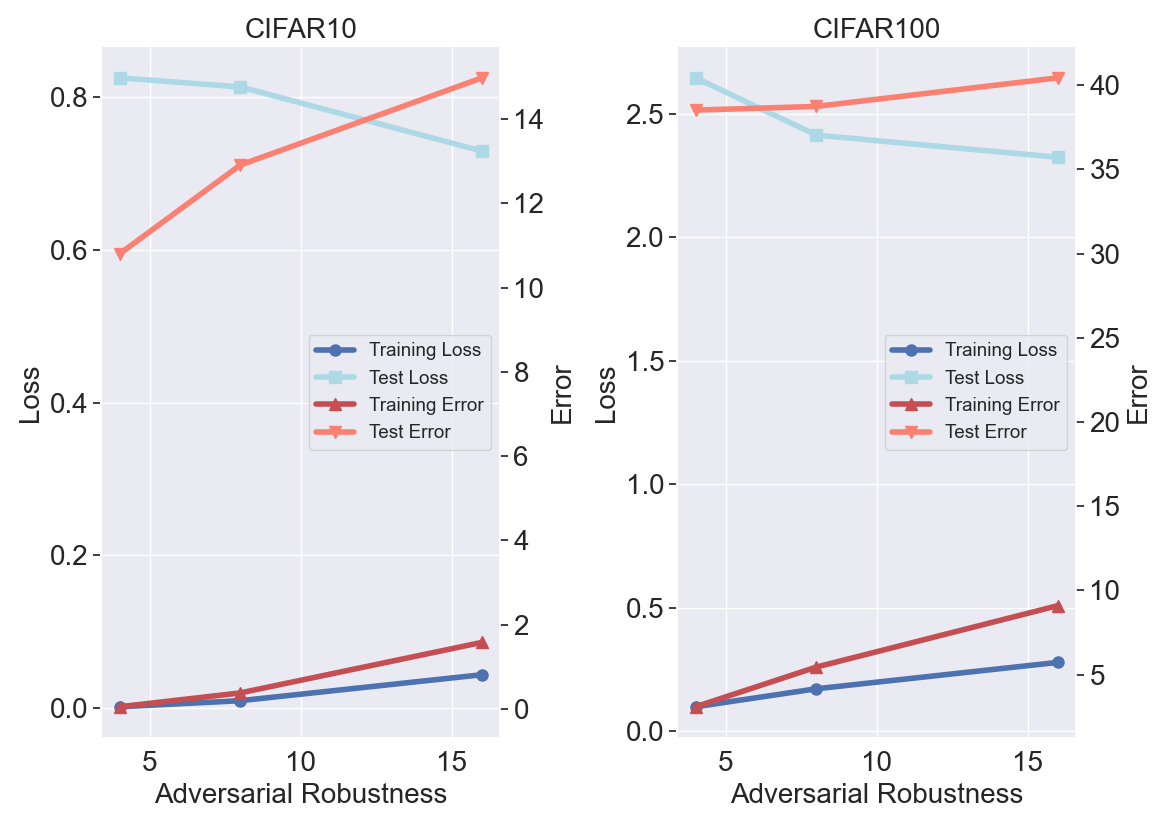}
      \caption{}
      \label{fig:performance_resnet110}
    \end{subfigure}
    \caption{
        Experiment results on CIFAR10/100 with ResNet-110 \citep{He2016}.  The
unit of x-axis is the adversarial robustness (AR) strength of NNs, c.f. the
beginning of \cref{sec:exper-valid}.  {\bf (a)} Plots of loss gap (and error
rate gap) between training and test datasets v.s. AR strength. {\bf (b)} Plots
of losses (and error rates) on training and test datasets v.s. AR strength.
        }
    \label{fig:datasets_resnet110}
\end{figure*}

To preliminarily validate that the regularization effects observed in
\cref{sec:regul-effects-nns} manifest in NNs with varied capacities, we
investigate the regularization effects of AR on a larger NNs, i.e., ResNet
110. The results are shown in \cref{fig:datasets_resnet110}. The observed
phenomenon is the same with that of ResNet56 presented in
\cref{sec:regul-effects-nns}, and thus corroborates our results.

\subsubsection{Regularization effects on NNs with controlled capacities}
\label{sec:regul-effects-nns-3}

To control capacities of NNs quantitatively, we choose the measure based on
spectral norm \citep{Bartlett2017,Neyshabur2017}. In spectral norm based
capacity measure bound \citep{Bartlett2017,Neyshabur2017}, the NN capacity is
normally proportional to a quantity called spectral complexity (SC), which is
defined as follows.

\begin{definition}[Spectral Complexity]
    Spectral complexity $\text{SC}(T)$ of a NN $T$ is the multiplication of spectral norms of weight matrices
    of layers in a NN.
    \begin{displaymath}
      \text{SC}(T) = \prod_{i=1}^{L}||\bv{W}_{i}||_2
    \end{displaymath}
    where $\{\bv{W}_{i}\}_{i=1\ldots L}$ denotes weight matrices of layers of the NN.
\end{definition}
To control SC, we apply the spectral normalization (SN) \citep{Sedghi2018a}
on NNs. The technique renormalizes the spectral norms of the weight matrices of
a NN to a designated value after certain iterations. We carry out the
normalization at the end of each epoch.

We train ResNet56 with increasingly strong AR and with
increasingly strong spectral normalization. The results are shown in
\cref{fig:control}.

As can be seen, as the capacity of NNs decreases (from upper left to bottom
right in each sub-figure), the error gap between
training and test gradually changes from an increasing trend to a decreasing
trend, while the loss gap keeps a consistent decreasing trend. It suggests that
the overfitting phenomenon is gradually prevented by another
regularization techniques, i.e., the spectral normalization. As a result, the
regularization effect of AR starts to emerge even in the error gap, which
previously manifests only in the loss gap.
The other curves corroborate our previous observations and analysis as
well.

\subsection{Further evidence on the smoothing effect of adversarial robustness}
\label{sec:furth-evid-smooth}

We quantitatively measure the smoothing effect around examples here by
measuring the average maximal loss change/variation induced by the perturbation (of a
fixed infinity norm) applied on examples. We found that the loss variation
decreases as networks become increasingly adversarially robust.  Note that the
loss of an example is a proxy to the confidence of the example --- it is the
logarithm of the estimated probability (a characterization of confidence) of
the NN classifier.

\begin{figure}[h]
  \centering
    \includegraphics[width=1.0\columnwidth]{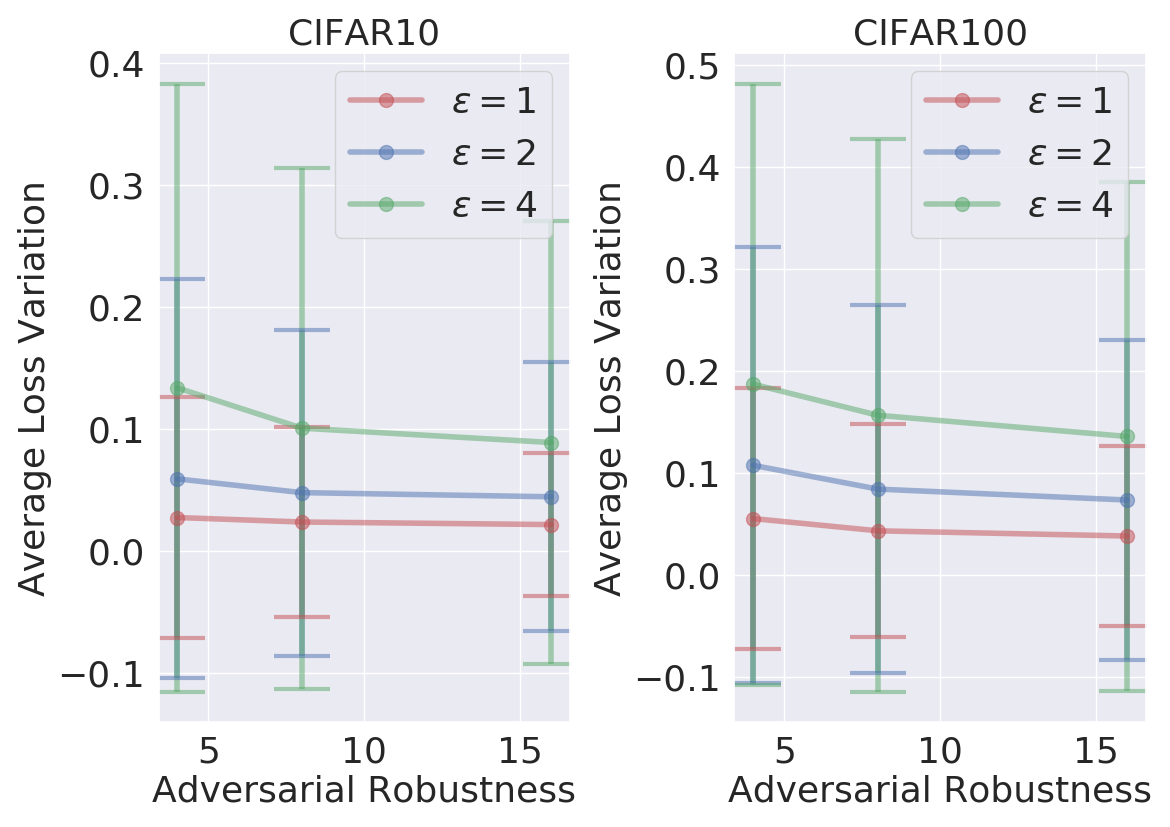}
  \caption{Average maximal loss variation induced by adversarial examples in
    networks with increasing adversarial robustness. The experiments are
    carried on CIFAR10/100. $\epsilon$ represents the maximal perturbation can be
applied on natural test examples to generate adversarial examples. It is
measured in the infinity norm.  The larger the $\epsilon$, the stronger the
perturbation is. The error bars represent standard
deviation.}
    \label{fig:loss_variation}
\end{figure}

For a given maximal perturbation range characterized by the infinity norm, we
generate adversarial examples within that norm for all test samples. For
each example, the maximal loss variation/change of the adversarial example
w.r.t. the natural example is computed for networks with different adversarial
strength. To obtain statistical behaviors, we compute the average and standard
deviation of such maxima of all test samples. The results are shown in
\cref{fig:loss_variation}. The exact data can be found in \cref{table:smooth_data}.

We can see that the average loss variation decreases with adversarial
robustness. The standard deviation decreases with network adversarial
robustness as well. The phenomenon that the standard deviation is comparably
large with the mean might need some explanation. This is because different
examples have different losses, thus the loss varies in relatively different
regimens --- the more wrongly classified examples vary in a larger magnitude, and
vice versa for more correctly classified examples. This phenomenon leads to
the large standard deviation of the loss variation.

\begin{table*}[h]
	\caption{
		Data of the smoothing effect of PGD adversarial training in \cref{fig:loss_variation}.
	}
        \label{table:smooth_data}
	\vskip 0.15in
	\begin{center}
		\begin{small}
			\begin{tabular}{c c c c c}
			\toprule
			\multirow{2}{*}{Dataset} & \multirow{2}{*}{Attack Strength} & \multicolumn{3}{c}{Defensive Strength} \\
			\cline{3-5}
			~ & ~ & $4$ & $8$ & $16$ \\
			\midrule
			\multirow{3}{*}{CIFAR10} & $\epsilon=1$ & $0.0273\pm0.0989$ & $0.0236\pm0.0778$ & $0.0215\pm0.0588$ \\
			~ & $\epsilon=2$ & $0.0590\pm0.1637$ & $0.0477\pm0.1337$ & $0.0443\pm0.1102$ \\
			~ & $\epsilon=4$ & $0.1337\pm0.2494$ & $0.1006\pm0.2137$ & $0.0888\pm0.1816$ \\
			\midrule
			\multirow{3}{*}{CIFAR100} & $\epsilon=1$ & $0.0550\pm0.1276$ & $0.0430\pm0.1043$ & $0.0379\pm0.0886$ \\
			~ & $\epsilon=2$ & $0.1072\pm0.2138$ & $0.0839\pm0.1802$ & $0.0732\pm0.1568$ \\
			~ & $\epsilon=4$ & $0.1868\pm0.2946$ & $0.1563\pm0.2712$ & $0.1355\pm0.2494$ \\
			\bottomrule
		\end{tabular}
	\end{small}
\end{center}
\vskip -0.1in
\end{table*}

\subsection{Further experiments on using FGSM in adversarial training to build adversarial robustness}
\label{sec:furth-empir-study}

We explain the choice of PGD as the representative of adversarial training
techniques here.  Various adversarial training methods are variant algorithms
that compute first order approximation to the point around the input example
that minimizes the label class confidence.  The difference is how close the
approximation is.  Recent works on adversarial examples exclusively only use
PGD in experiments
\citep{kannan2018adversarial,Schmidt,xie2019feature,ilyas2019adversarial,wang2019bilateral}.
It is also a very strong multi-step attack method that improves over many of
its antecedents: NNs trained by FGSM could have no defense ability to
adversarial examples generated by PGD, as shown in Table 5 in
\citet{Madry2017}; multi-step methods prevent the pitfalls of adversarial
training with single-step methods that admit a degenerate global minimum
\citep{tramer2017ensemble}.  Thus, we believe the observations in this work is
representative for various adversarial training techniques.  Yet, even in the
worst case, this work at least makes a first step to understand a
representative approach of the approximation.

To corroborate the analysis, we also use FGSM \citep{Goodfellow2015a} in the adversarial training to build adversarial robustness into NNs. 
The results are consistent with the results obtained using PGD.  
The experiments are carried on CIFAR10/100. 
We present key plots that support the results obtained in the main con- tent here. 
All the setting are same with that described in \cref{sec:techn-build-advers} of PGD, except that we replace PGD with FGSM.

\begin{figure}[h]
  \centering
  \begin{subfigure}[b]{0.40\textwidth}
    \includegraphics[width=\linewidth]{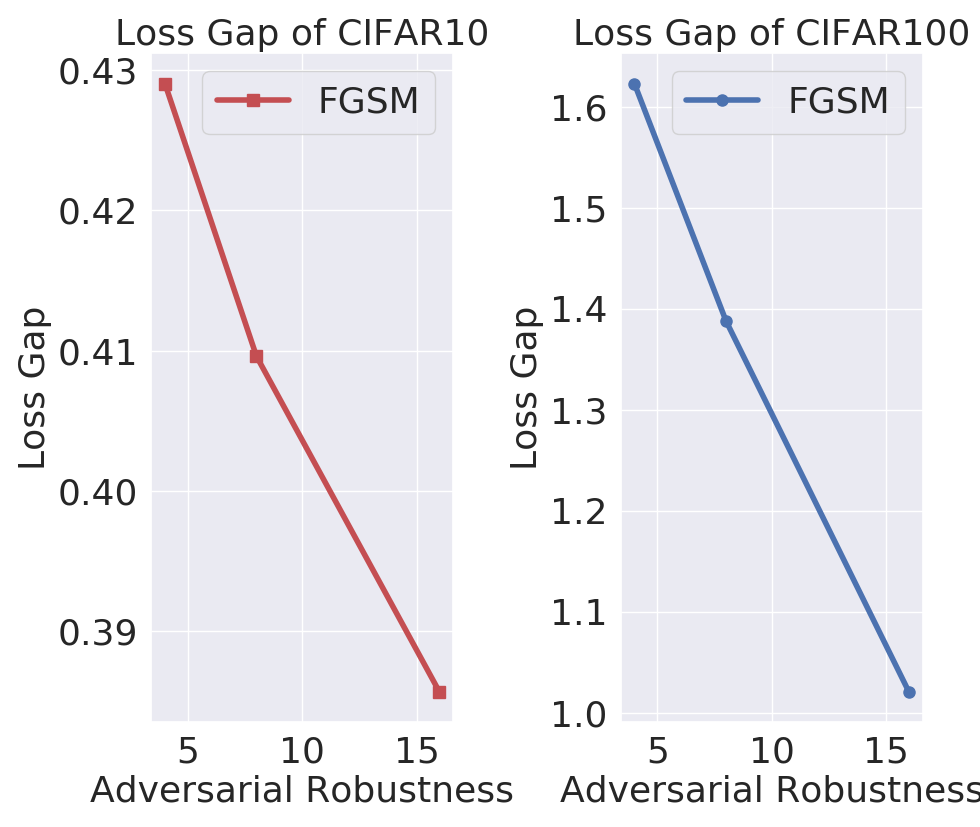}
    \caption{}
    \label{fig:loss_gap_fgsm}
  \end{subfigure}
  \\
  \begin{subfigure}[b]{0.40\textwidth}
    \includegraphics[width=\linewidth]{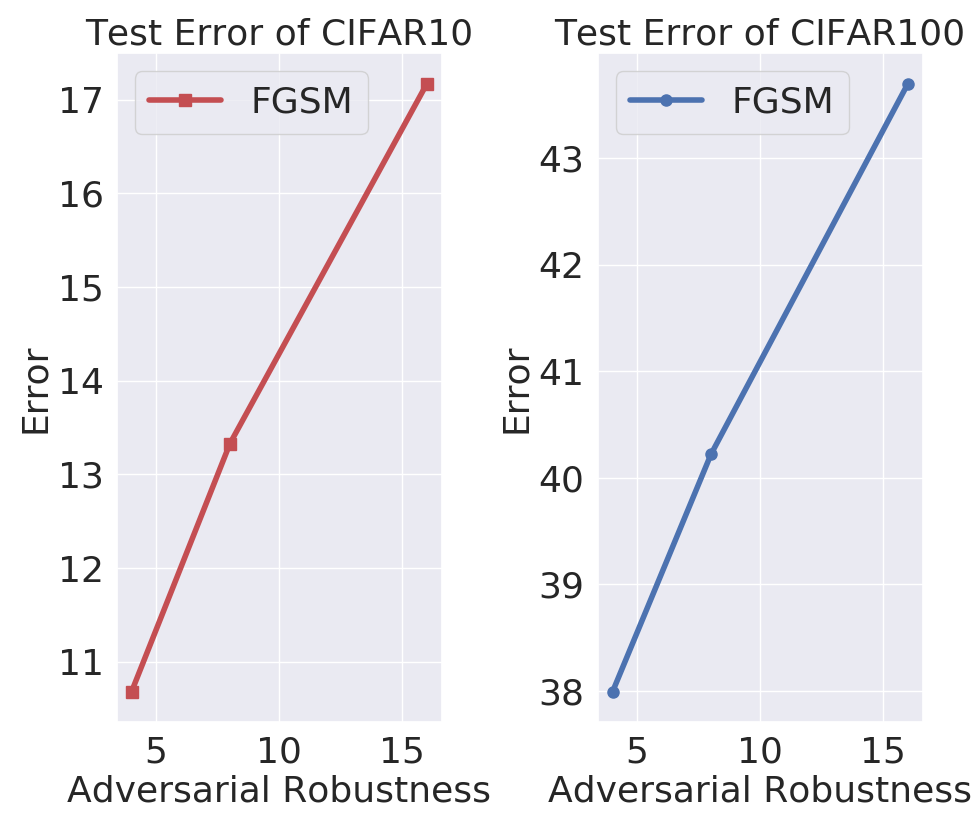}
    \caption{}
    \label{fig:performance_fgsm}
  \end{subfigure}
  \caption{
    Experiment results on CIFAR10/100. 
    The network is ResNet-56 \citep{He2016}. The unit of x-axis is the adversarial robustness (AR) strength of NNs, c.f. the beginning of \cref{sec:exper-valid}.
    {\bf (a)} Plots of loss gap between training and test datasets v.s. AR strength. 
    {\bf (b)} Plots of error rates on training and test datasets v.s. AR strength.}
  \label{fig:datasets_fgsm}
\end{figure}

\paragraph{Adversarial robustness reduces generalization gap and standard test performance.}
In \cref{sec:regul-effects-nns}, we find that NNs with stronger adversarial
robustness tend to have smaller loss/generalization gap between training and
test sets. Consistent phenomenon has been observed in networks adversarially
trained with FGSM on CIFAR10/100, as shown in
\cref{fig:loss_gap_fgsm}. Consistent standard test performance degradation has
been observed in adversarially trained with FGSM on CIFAR10/100 as well, as shown in
\cref{fig:performance_fgsm}. The exact data can be found in
\cref{table:gsfm_data}.

\begin{figure}[h]
  \centering
  \begin{subfigure}[b]{0.235\textwidth}
    \includegraphics[width=\columnwidth]{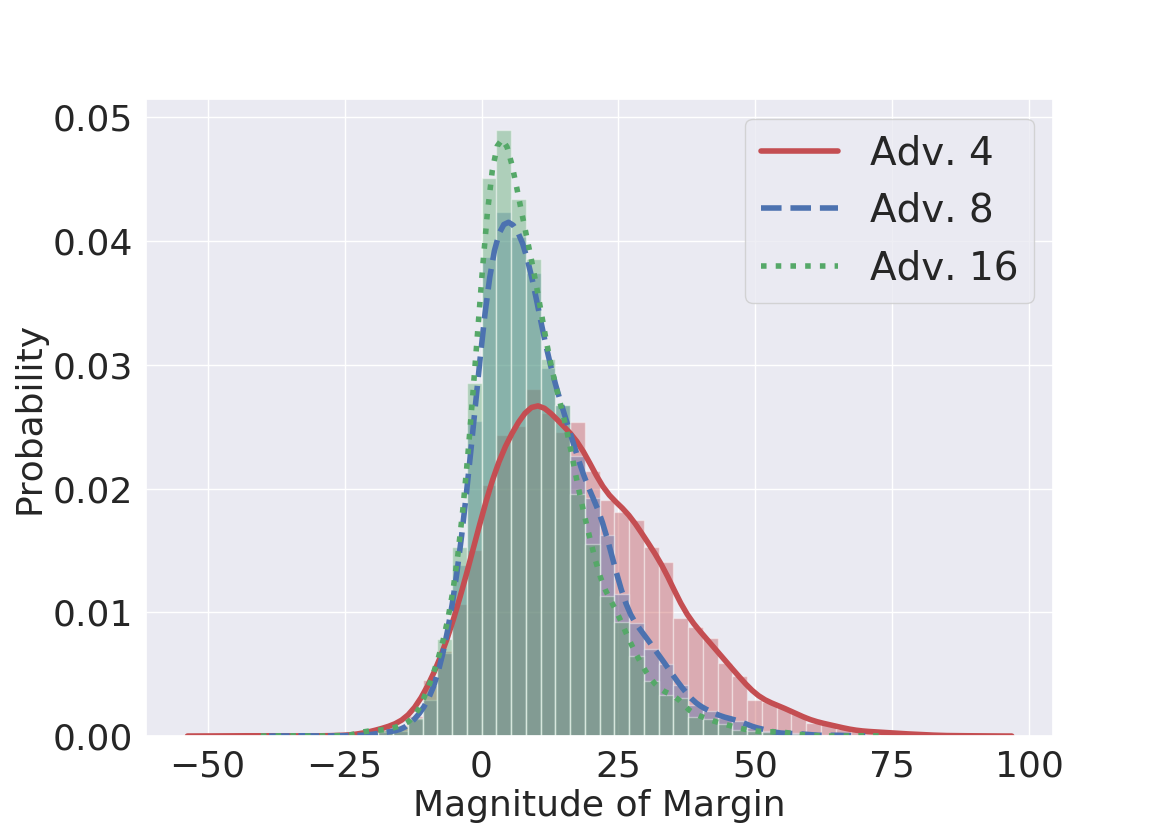}
    \caption{CIFAR10 Test}
    \label{fig:cifar10_margin_test_fgsm}
  \end{subfigure}
  \begin{subfigure}[b]{0.235\textwidth}
    \includegraphics[width=\columnwidth]{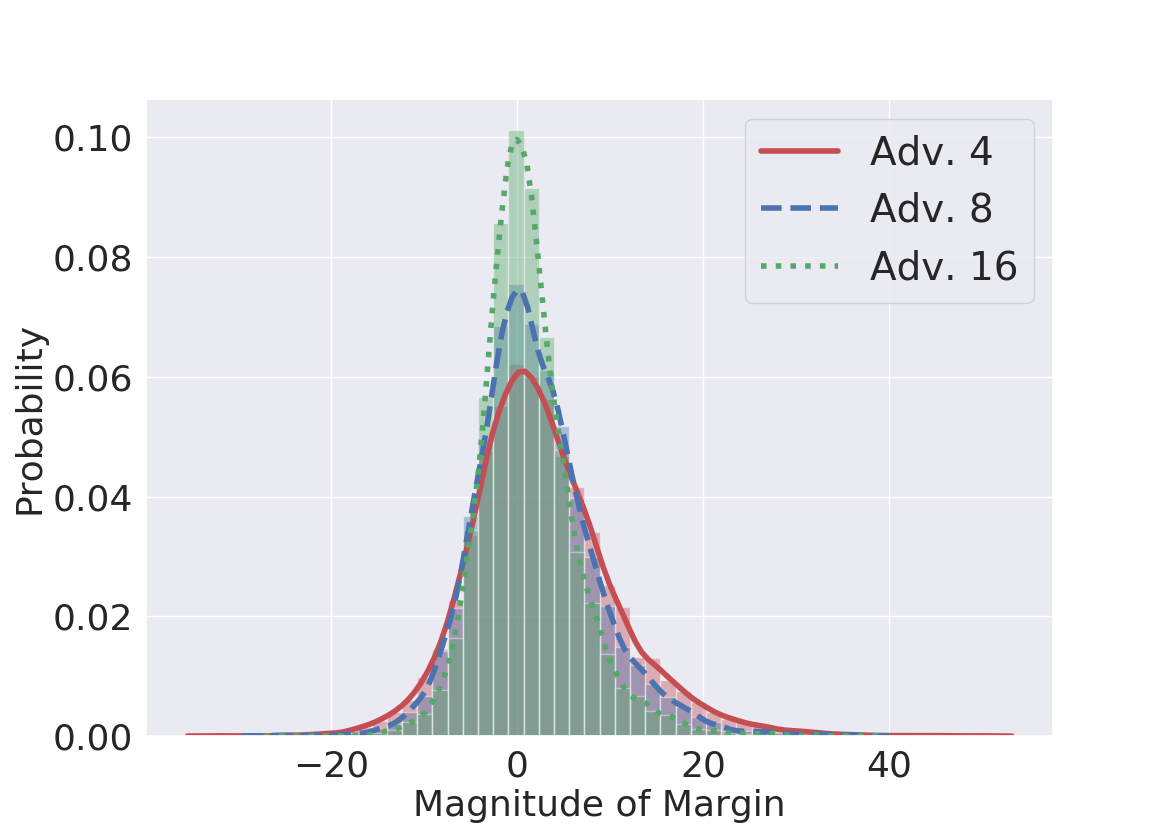}
    \caption{CIFAR100 Test}
    \label{fig:cifar100_margin_test_fgsm}
  \end{subfigure}
  \\
  \begin{subfigure}[b]{0.235\textwidth}
    \includegraphics[width=\columnwidth]{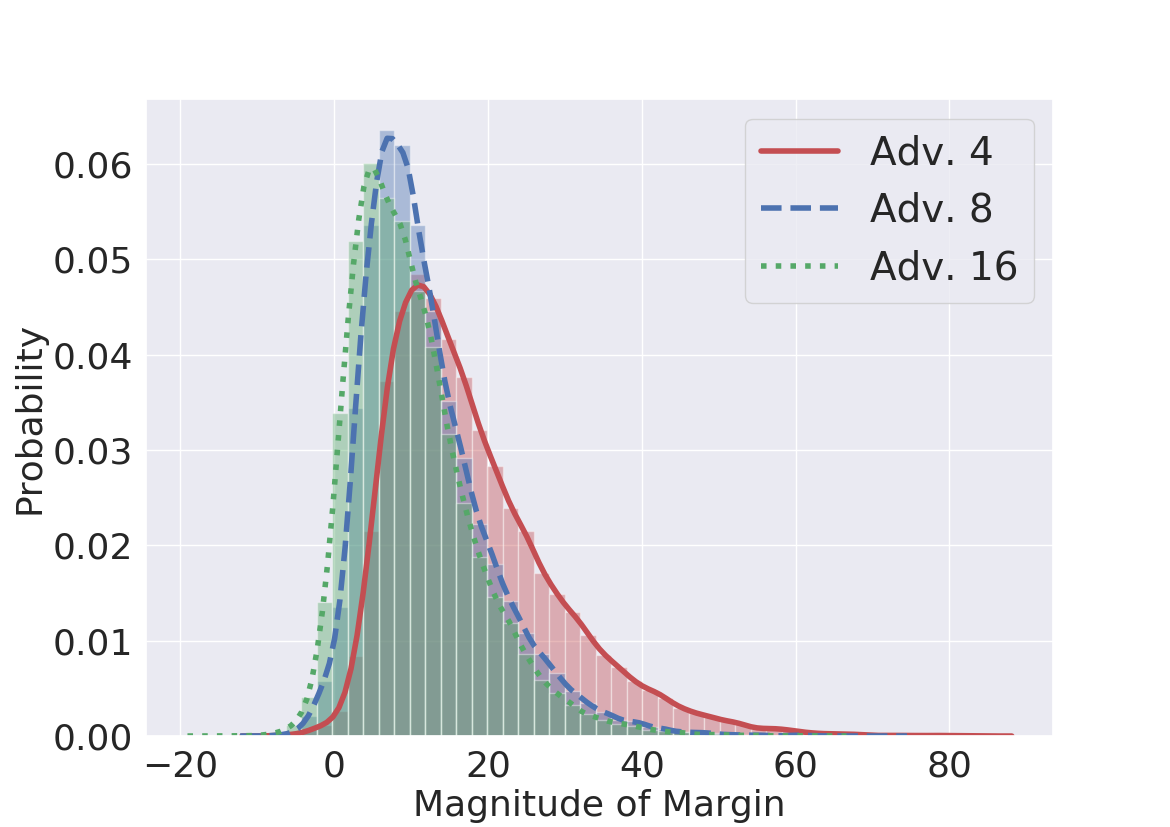}
    \caption{CIFAR10 Training}
    \label{fig:cifar10_margin_train_fgsm}
  \end{subfigure}
  \begin{subfigure}[b]{0.235\textwidth}
    \includegraphics[width=\columnwidth]{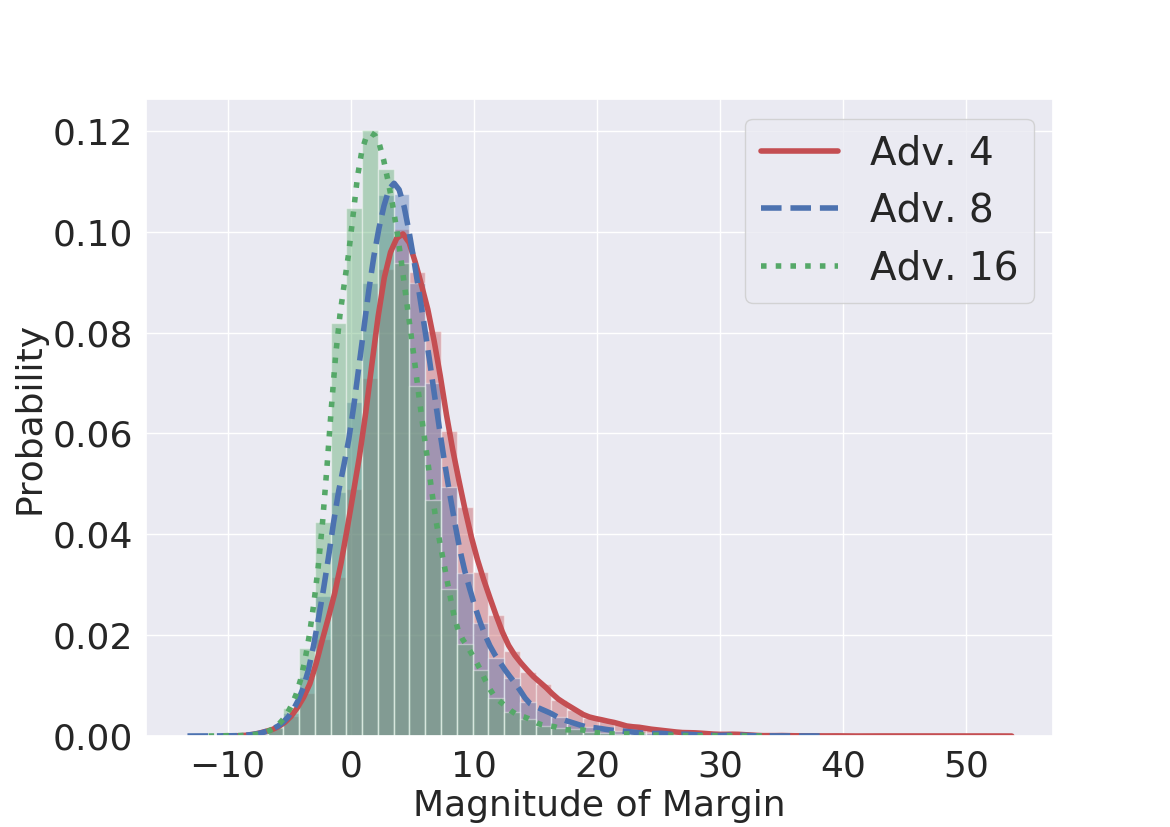}
    \caption{CIFAR100 Training}
    \label{fig:cifar100_margin_train_fgsm}
  \end{subfigure}
  \caption{Margin distributions of NNs with AR strength 4, 8, 16 on Training and
    Test sets of CIFAR10/100.}
    \label{fig:margin_distribution_fgsm}
\end{figure}

\begin{figure}[h]
  \centering
  \begin{subfigure}[b]{0.235\textwidth}
    \includegraphics[width=\columnwidth]{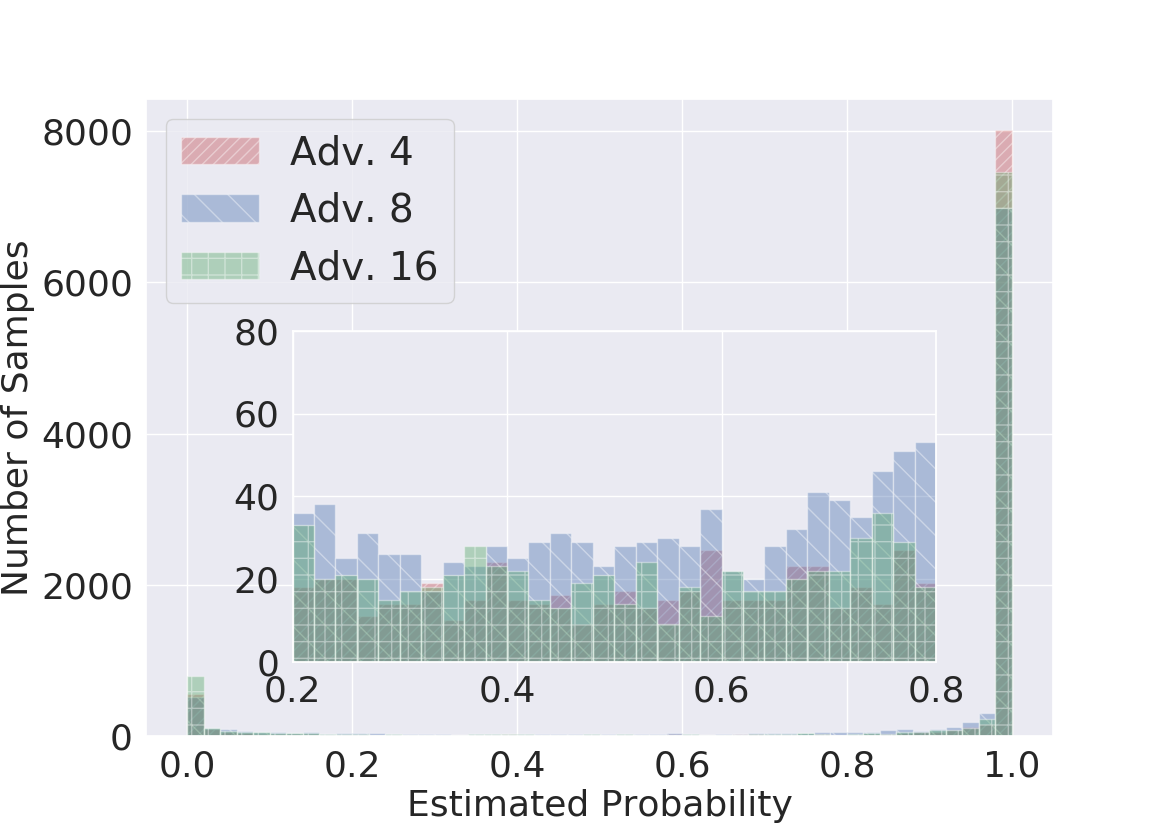}
    \caption{Prob. Histogram}
    \label{fig:prob_histogram_fgsm}
  \end{subfigure}
  \begin{subfigure}[b]{0.235\textwidth}
    \includegraphics[width=\columnwidth]{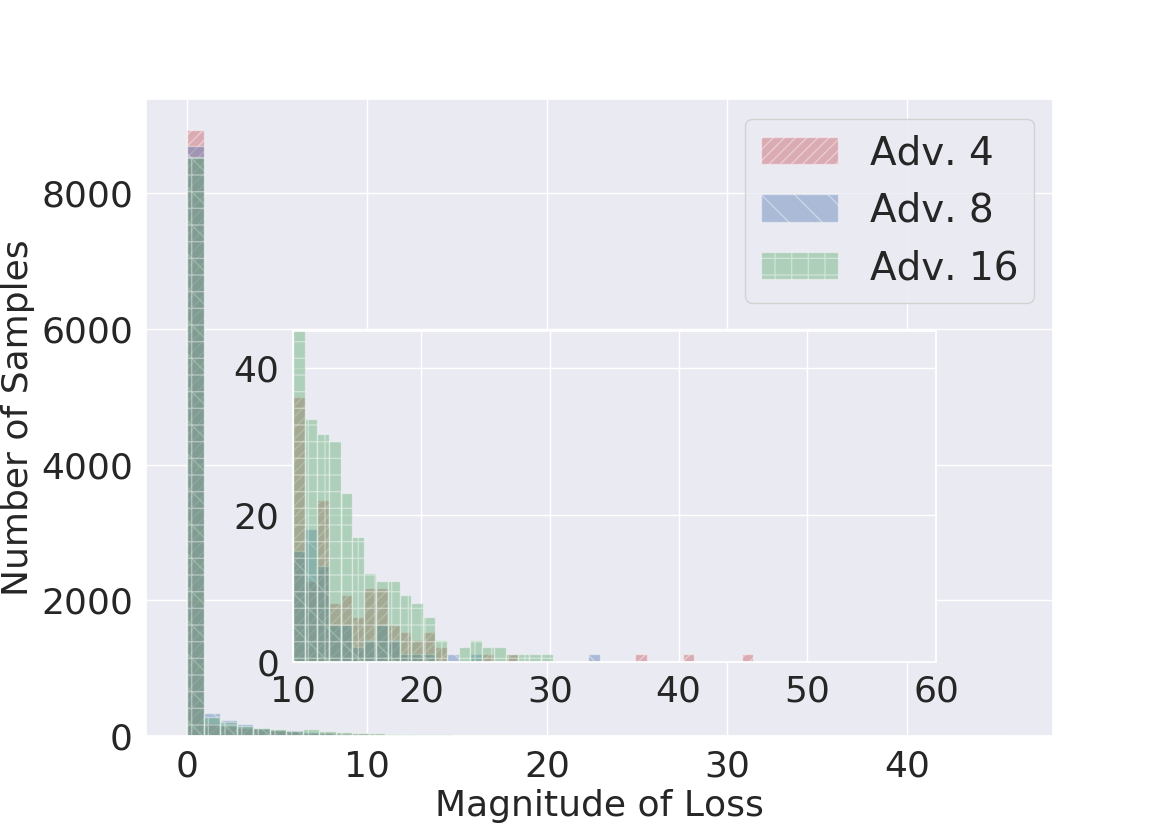}
    \caption{Loss Histogram}
    \label{fig:loss_histogram_fgsm}
  \end{subfigure}
  \\
  \begin{subfigure}[b]{0.235\textwidth}
    \includegraphics[width=\columnwidth]{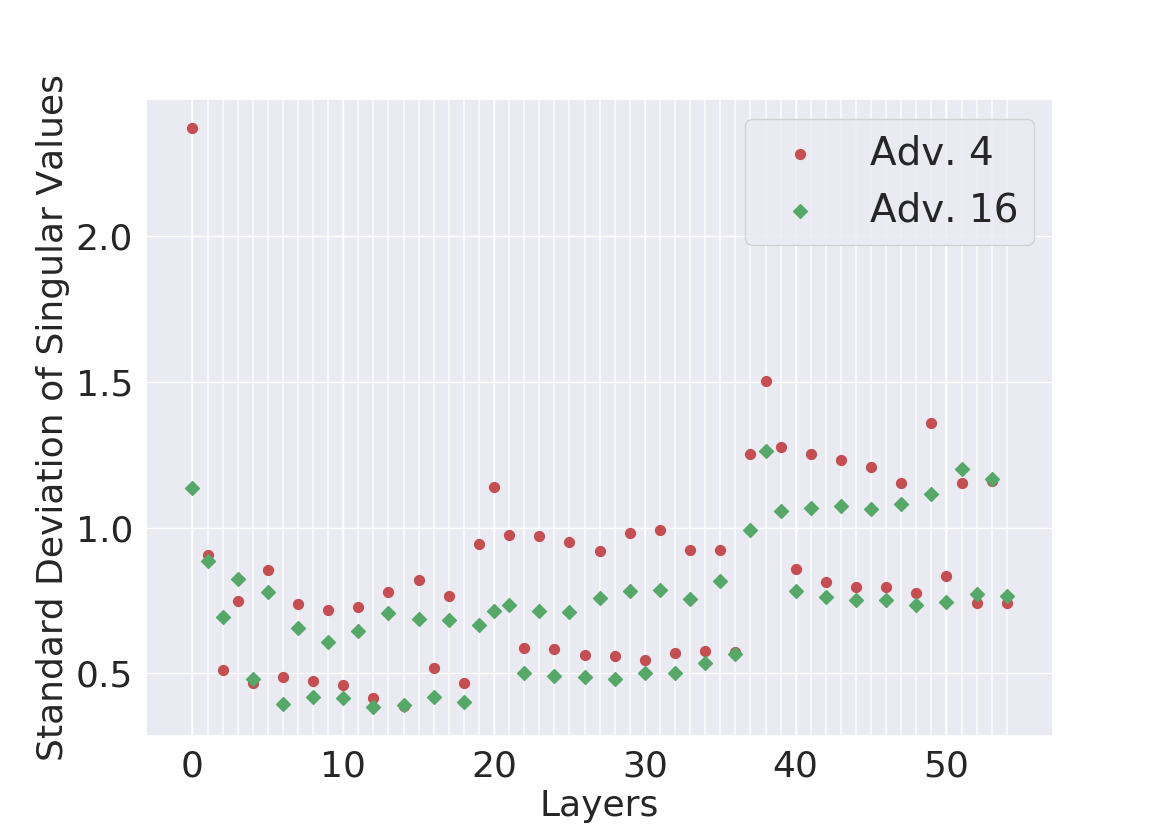}
    \caption{CIFAR10}
    \label{fig:singular_value_CIFAR10_fsgm}
  \end{subfigure}
  \begin{subfigure}[b]{0.235\textwidth}
    \includegraphics[width=\columnwidth]{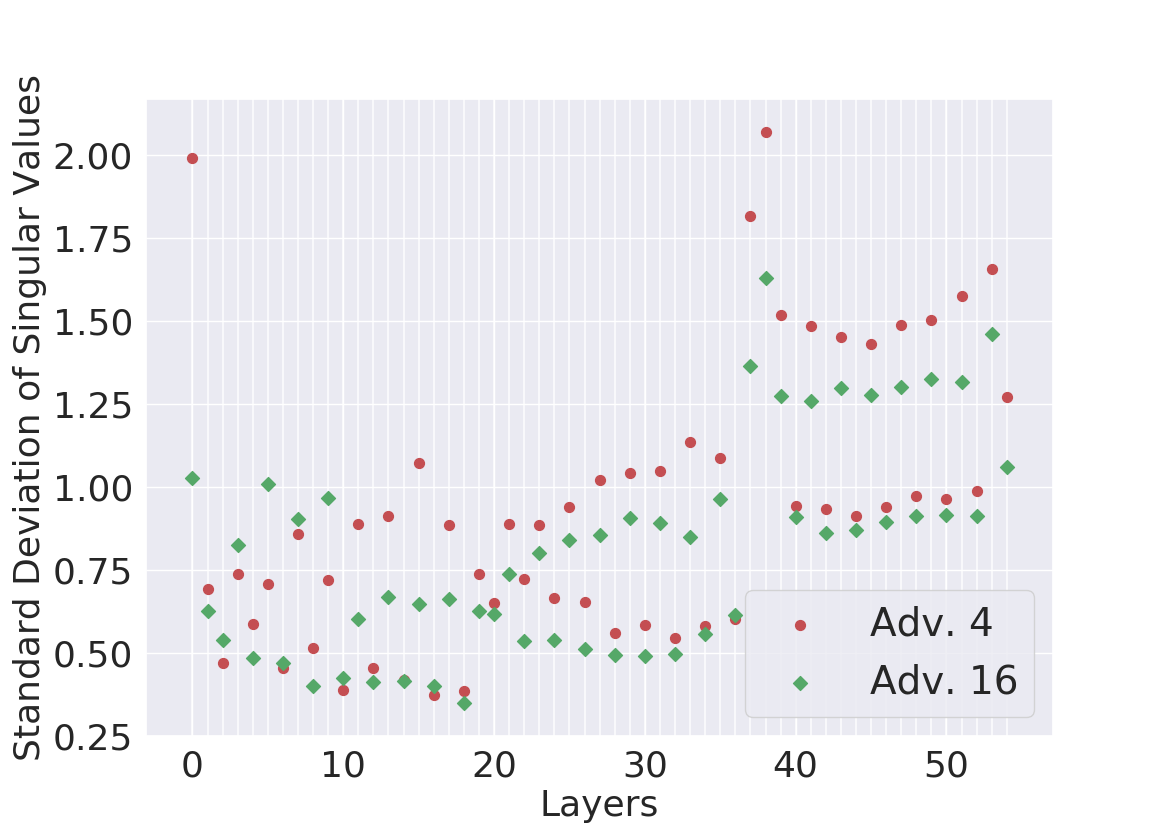}
    \caption{CIFAR100}
    \label{fig:singular_value_CIFAR100_fsgm}
  \end{subfigure}
  \caption{{\bf (a)(b)} are histograms of estimated probabilities and losses
    respectively of the test
    set sample of NNs trained AR strength 4, 8, 16. We plot a subplot of a narrower range inside the plot of
the full range to show the histograms of examples that are around the middle
values to show the change induced by AR that induces more middle valued
confidence predictions. {\bf (c)(d)} are standard deviations of singular values of
weight matrices of NNs at each layer trained on CIFAR10/100 with AR strength 4,
16. The AR strength 8 is dropped for clarity.}
    \label{fig:misc_fgsm}
\end{figure}

\paragraph{Adversarial robustness concentrates examples around decision boundaries.}
In \cref{sec:cms-that-concentrate}, we find that the distributions of margins become more concentrated around zero as AR grows. 
The phenomenon has been observed consistently in networks adversarially trained with FGSM on CIFAR10/100, as shown in \cref{fig:margin_distribution_fgsm}. 
Phenomenon in \cref{fig:prob_histogram} and \cref{fig:loss_histogram} are also reproduced consistently in \cref{fig:prob_histogram_fgsm} and \cref{fig:loss_histogram_fgsm}. 
Please refer to \cref{sec:cms-that-concentrate} for the analysis of the results. 
Here we mainly present counterparts of the results analyzed there.

\paragraph{Adversarial robustness reduces the standard deviation of singular
  values of weight matrices in the network.}
In \cref{sec:advers-robustn-makes}, we find that for NNs with stronger adversarial robustness, the standard deviation of singular values of weight matrices is smaller in most layers. The phenomenon has been consistently observed in NNs trained with FGSM on CIFAR10/100, as shown in \cref{fig:singular_value_CIFAR10_fsgm} and
\cref{fig:singular_value_CIFAR100_fsgm}. 
Please refer to \cref{sec:advers-robusnt-leads} and \cref{sec:advers-robustn-makes} for the analysis of the results. 
Here we mainly present counterparts of the results analyzed there.

In conclusion, all key empirical results have been consistently observed in NNs
trained with FGSM.

\begin{table}[h]
\caption{
 Data of \cref{fig:datasets_fgsm}
}
\label{table:gsfm_data}
\vskip 0.15in
\begin{center}
	\begin{small}
		\begin{tabular}{c c c c c}
			\toprule
			\multirow{2}{*}{Dataset} & ~ & \multicolumn{3}{c}{Defensive Strength} \\
			~ & ~ & $4$ & $8$ & $16$ \\
			\midrule
			\multirow{5}{*}{CIFAR10} & Test Acc. & $89.32$ & $86.67$ & $82.83$ \\
			\cline{2-5}
			~ & Trn Loss & $0.038$ & $0.086$ & $0.252$ \\
			~ & Test Loss & $0.467$ & $0.495$ & $0.637$ \\
			\cline{2-5}
			~ & $\Delta$ Loss & $0.429$ & $0.409$ & $0.385$ \\
			\midrule
			\multirow{5}{*}{CIFAR100} & Test Acc. & $62.01$ & $59.78$ & $56.30$ \\
			\cline{2-5}
			~ &  Trn Loss & $0.469$ & $0.656$ & $0.822$ \\
			~ & Test Loss & $1.776$ & $1.723$ & $1.797$ \\
			\cline{2-5}
			~ & $\Delta$ Loss & $1.307$ & $1.067$ & $0.975$ \\
			\bottomrule
		\end{tabular}
	\end{small}
\end{center}
\vskip -0.1in
\end{table}

\clearpage
%\pagebreak
%-------------------------------------------------------------------------
\section{Proof of theorem 3.1}
\label{sec:proof-}

\subsection{Algorithmic Robustness Framework}
\label{sec:robustness}

In order to characterize the bound to the GE, we build on the {\it algorithmic robustness} framework \citep{Xu2012a}.

We introduce the framework below.

\begin{definition}[$(K, \epsilon(\cdot))$-robust]
\label{def:robust}
An algorithm is $(K, \epsilon(\cdot))$ robust, for $K \in \mathbb{N}$ and $\epsilon(\cdot):
\mathcal{Z}^{m} \mapsto \mathbb{R}$, if $\mathcal{Z}$ can be partitioned into $K$
disjoint sets, denoted by $\mathcal{C} = \{ C_k \}_{k=1}^{K}$, such that the following
holds for all $s_i = (\bvec{x}_i, y_i) \in S_m, z=(\bvec{x}, y) \in \mathcal{Z}, C_k \in \mathcal{C}$:
\begin{align*}
    \forall s_i = (\bvec{x}_i, y_i) \in C_k, \forall z=(\bvec{x}, y) \in C_k \\
    \implies |l(f(\bvec{x}_i), y_i) - l(f(\bvec{x}), y)| \leq \epsilon(S_m).
\end{align*}
\end{definition}
The gist of the definition is to constrain the variation of loss values on test
examples w.r.t. those of training ones through local property of the
algorithmically learned function $f$. Intuitively, if $s \in S_m$ and $z
\in \mathcal{Z}$ are ``close'' (e.g., in the same partition $C_k$), their loss
should also be close, due to the intrinsic constraint imposed by
$f$.

For any algorithm that is robust, Xu \& Mannor
\citep{Xu2012a} proves
\begin{theorem}[Xu \& Mannor \citep{Xu2012a}]
\label{thm:1}
If a learning algorithm is $(K, \epsilon(\cdot))$-robust and $\mathcal{L}$ is bounded,
a.k.a. $\mathcal{L}(f(\bvec{x}), y) \leq M$ $\forall z \in \mathcal{Z}$, for any $\eta > 0$, with probability
at least $1 - \eta$ we have
\begin{equation}
    \label{abs_robust_bound}
    \GE(f_{S_m}) \leq \epsilon(S_m) + M\sqrt{\frac{2K\log(2) + 2 \log(1/\eta)}{m}} .
\end{equation}
\end{theorem}
To control the first term, an approach is to constrain the variation of the
loss function.  Covering number (Shalev-Shwartz \& Ben-David,
\citep{shalev2014understanding}, Chapter 27) provides a way to characterize
the variation of the loss function, and conceptually realizes the actual number
$K$ of disjoint partitions.

For any regular $k$-dimensional manifold embedded in space equipped with a
metric $\rho$, e.g., the image data embedded in $L^2(\bb{R}^{2})$, the square integrable function
space defined on $\bb{R}^{2}$, it has a covering number $\ca{N}(\ca{X}; \rho,
\epsilon)$ of $(C_{\ca{X}}/\epsilon)^{k}$ \citep{Verma2013}, where $C_{\ca{X}}$ is a
constant that captures its ``intrinsic'' properties, and $\epsilon$ is the radius
of the covering ball. When we calculate the GE bound of NNs, we would assume
the data space is a $k$-dimensional regular manifold that accepts a covering.

Adversarial robustness makes NNs a $(K, \epsilon(\cdot))$-robust algorithm, and is able to control the variation of loss values on test examples. 
Building on covering number and \cref{thm:1}, we are able to prove \cref{thm:main}.

\subsection{Proof}
\label{sec:proof}

\begin{proof}[{\bf Proof of \cref{lm:1}
    }]
    By theorem 3 in \citet{Sokolic2017}, we have
    \begin{equation}
      \label{eq:fmap_diff}
      \left|\left|  I_{l}(\bv{x}) - I_{l}(\bv{x}') \right|\right|
      = \left|\left| \int_{0}^{1}\bv{J}(\bv{x} - t(\bv{x}' - \bv{x}))dt(\bv{x} - \bv{x}') \right|\right|\\
    \end{equation}
    where $\bv{J}(\bv{x})$ denotes the Jacobian of $I_l(\bv{x})$ at $\bv{x}$.
  
    By lemma 3.2 in \citet{Jia2019}, when we only have max pooling layers and ReLU
    as nonlinear layer in NNs, $\bv{J}(\bv{x})$ is a linear operator at a local
    region around $\bv{x}$. For terminology concerning regions, we follow the definitions in
    \citet{Jia2019}. More specifically, we have
    \begin{displaymath}
      \bv{J}(\bv{x}) = \prod_{i=1}^{l}\bv{W}_i^{\bv{x}}
    \end{displaymath}
    where $\bv{W}_{i}^{\bv{x}}$ is the linear mapping (matrix) induced by
    $\bv{J}(\bv{x})$ at $\bv{x}$. It is a matrix obtained by selectively setting
    certain rows of $\bv{W}_i$ to zero. For the more concrete form of $\bv{W}_i^{\bv{x}}$,
    refer to lemma 3.2 in \citet{Jia2019}. In \citet{Jia2019}, it is noted as
    $\bv{W}^{q}_i$, where $q$ is a region where $\bv{x}$ is in.
  
    Suppose that from $\bv{x}$ to $\bv{x}'$, the line segment $\bv{x} - \bv{x}'$ passes through regions $\{q_j\}_{j=1,\ldots, n}$. 
    The line segment is illustrated in \cref{fig:piecewise} as the boldest black line segment at the upper half of the figure. 
    In the illustration, $\bv{x} - \bv{x}'$ passes through three regions, colored coded as gray, dark yellow, light blue respectively. 
    The line segment is divided into three sub-segments. 
    Suppose $\bv{l}(t) = \bv{x} + t(\bv{x}' - \bv{x})$. 
    Then the three sub-segments can be represented by $\bv{l}(t)$ as $\bv{l}(s_1)$ to $\bv{l}(e_1)$, $\bv{l}(s_2)$ to $\bv{l}(e_2)$, and $\bv{l}(s_3)$ to $\bv{l}(e_3)$ respectively, as noted on the line segment in the illustration. 
    Originally, the range of the integration in \cref{eq:fmap_diff} is from $0$ to $1$, representing the integration on the line segment $\bv{l}(0)$ to $\bv{l}(1)$ in the instance space. 
    Now, since for each of these regions trespassed by the line segment, the Jacobian $\bv{J}(\bv{x})$ is a linear operator, denoted as $\bv{W}^{q_j}_i$, the integration in \cref{eq:fmap_diff} from $0$ to $1$ can be decomposed as a summation of integration on segments $l(s_1)$ to $l(e_1)$ etc. In each of these integration, the Jacobian $\bv{J}(\bv{x})$ is the multiplication of linear matrices $\bv{W}^{q_j}_i$, i.e., $\prod_{i=1}^{l}\bv{W}_{i}^{q_j}$.
    Thus, \cref{eq:fmap_diff} can be written as
    \begin{displaymath}
      \sum_{j=1}^{n}\int_{s_j}^{e_j}\left|\left|\prod_{i=1}^{l}\bv{W}_{i}^{q_j}dt(\bv{x} - \bv{x}')\right|\right|
    \end{displaymath}
  where $s_j, e_j$ denotes the start and end of the segment $[s_j, e_j] \subset [0,
    1]$ of the segment $[0, 1]$ that passes through the region $q_j$.
  
\end{proof}

In the cases that a linear operator is applied on the feature map
$I_{l}(\bv{x})$ without any activation function, we can also obtain a similar
conclusion. Actually, such cases are just degenerated cases of feature maps
that have activation functions.
\begin{corollary}
  \label{cl:1}
  Given two elements $\bv{x}, \bv{x}'$, and $I_l(\bv{x}) =
  \bv{W}_lg(\bv{W}_{l-1}\ldots g(\bv{W}_1\bv{x}))$, we have
\begin{displaymath}
  \left|\left|  I_{l}(\bv{x}) - I_{l}(\bv{x}') \right|\right| =
    \sum_{j=1}^{n}\int_{s_j}^{e_j}\left|\left|\bv{W}_l\prod_{i=1}^{l-1}\bv{W}_{i}^{q_j}dt(\bv{x} - \bv{x}')\right|\right|
\end{displaymath}
where symbols are defined similar as in Proof of \cref{lm:1}.
\end{corollary}

Now, we are ready to prove \cref{thm:1}.

\begin{proof}[{\bf Proof of \cref{thm:1}}]
  Similar with the proof of \cref{thm:1}, we partition space $\mathcal{Z}$ into
the $\epsilon$-cover of $\ca{Z}$, which by assumption is a $k$-dimension manifold.
Its covering number is upper bounded by $C_{\mathcal{X}}^{k}/\epsilon^{k}$,
denoting $K = C_{\mathcal{X}}^{k}/\epsilon^{k}$, and $\hat{C}_i$ the $i$th covering
ball. For how the covering ball is obtained from the $\epsilon$-cover, refer to
theorem 6 in \citet{Xu2012a}.
We study the constraint/regularization that adversarial robustness
imposes on the variation of the loss function. Since we only have
$\epsilon$-adversarial robustness, the radius of the covering balls is at most
$\epsilon$ --- this is why we use the same symbol. Beyond $\epsilon$, adversarial
robustness does not give information on the possible variation anymore.
Let $T'$ denotes the NN without the last layer.

First, we analyze the risk change in a covering ball $C_i$. The analysis is divided
into two cases: 1 all training samples in $C_i$ are classified
correctly; 2) all training samples in $C_i$ are classified wrong. Note that no
other cases exist, for that the radius of $C_i$ is restricted to be
$\epsilon$, and we work on $\epsilon$-adversarial robust classifiers. It guarantees that all samples in a ball are classified as the same
class. Thus, either all training samples are all classified correctly, or wrongly.

We first study case 1). Given any example $z = (\bv{x}, y) \in C_i$, let
$\hat{y} = \argmax_{i\not= y} \bv{w}^{T}_{i}T'\bv{x}$. Its
ramp loss is
\begin{displaymath}
l_{\gamma}(\bv{x}, y) = \max \{0, 1 - \frac{1}{\gamma} (\bv{w}_{y} - \bv{w}_{\hat{y}})^{T}T'\bv{x}\}.
\end{displaymath}
Note that within $C_i$, $(\bv{w}_{y} - \bv{w}_{\hat{y}})^{T}T'\bv{x} \geq 0$, thus $l_\gamma(\bv{x}, y)$ is mostly $1$, and we would not reach the region where $r > 0$ in \cref{def:ramp_loss}.
Let $u(\bv{x}) := (\bv{w}_{y} - \bv{w}_{\hat{y}})^{T}T'\bv{x}$, and $u^i_{\min} =
\min_{\forall \bv{x} \in C_i } u(\bv{x})$. We have
\begin{displaymath}
l_{\gamma}(\bv{x}, y) \leq \max \{0, 1 - \frac{u^i_{\min}}{\gamma}\} \leq \max \{0, 1 - \frac{u_{\min}}{\gamma}\},
\end{displaymath}
where $u_{\min}$ denotes the smallest margin
among all partitions.

The inequality above shows adversarial robustness requires that $T'\bv{x}$ should vary slowly
enough, so that in the worst case, the loss variation within the adversarial
radius should satisfy the above inequality. The observation leads to the
constraint on the loss difference $\epsilon(\cdot)$ defined earlier in
\cref{def:robust} in the following.

Given any training example $z := (\bv{x}, y) \in C_i$, and any element $z' := (\bv{x}', y') \in
C_i$, where $C_i$ is the covering ball that covers $\bv{x}$, we have
  \begin{align}
   & |l_{\gamma}(\bv{x}, y) - l_{\gamma}(\bv{x}', y')|                                    \nonumber \\
 = & |\max\{0, 1 - \frac{u(\bv{x})}{\gamma}\} - \max\{0, 1 - \frac{u(\bv{x}')}{\gamma}\}| \nonumber \\
 %% \leq & \max\{0, 1 - \frac{u(\bv{x})}{\gamma}\} + \max\{0, 1 - \frac{u(\bv{x}')}{\gamma}\} \nonumber \\
 %% \leq & 2\max\{0, 1 - \frac{u_{\min}}{\gamma}\} \nonumber\\
 \leq & \max\{0, 1 - \frac{u_{\min}}{\gamma}\} \label{eq:l_right_case}.
  \end{align}

Now we relate the margin to the margin in the instance
space.

Given $z := (\bv{x}, y) \in \ca{Z}$, and $z'$, of which $\bv{x}'$ is the
closest points to $\bv{x}$ (measured in Euclidean norm) on the decision
boundary, we can derive the inequality below.
\begin{align}
u(\bv{x}) &= u(\bv{x}) - u(\bv{x}')\\
&= \int_{0}^{1}\bv{J}(\bv{x} - t(\bv{x} - \bv{x}'))dt(\bv{x} - \bv{x}')\label{eq:Jacobian}\\
&= \int_{0}^{1}(\bv{w}_{y} - \bv{w}_{\hat{y}})^{T}\prod_{i=1}^{L-1}\bv{W}_{i}^{\bv{x} - t(\bv{x} - \bv{x}')}dt(\bv{x} - \bv{x}')\nonumber\\
&= \int_{0}^{1}\left|(\bv{w}_{y} - \bv{w}_{\hat{y}})^{T}\prod_{i=1}^{L-1}\bv{W}_{i}^{\bv{x} - t(\bv{x} - \bv{x}')}dt(\bv{x} - \bv{x}')\right|\label{eq:positive_score}\\
&=\sum_{j=1}^{n}\int_{s_j}^{e_j}\left|(\bv{w}_{y} - \bv{w}_{\hat{y}})^{T}\prod_{i=1}^{L-1}\bv{W}_{i}^{q_j}dt(\bv{x} - \bv{x}')\right|\label{eq:segment_integral}\\
&\geq \min_{y,\hat{y} \in \ca{Y}, y\not=\hat{y}}||\bv{w}_{y} - \bv{w}_{\hat{y}}||_2\prod_{i=1}^{L-1}\sigma_{\min}^{i}||\bv{x} -
    \bv{x}'||_2\int_{0}^{1}dt \label{eq:contraction_lower_bound}\\
&\geq \min_{y,\hat{y} \in \ca{Y}, y\not=\hat{y}}||\bv{w}_{y} - \bv{w}_{\hat{y}}||_2\prod_{i=1}^{L-1}\sigma_{\min}^{i}||\bv{x} - \bv{x}'||_2\nonumber
\end{align}
where $\bv{J}(\bv{x})$ denotes the Jacobian of $I_l(\bv{x})$ at $\bv{x}$.
\cref{eq:Jacobian} can be reached by theorem 3 in \citet{Sokolic2017}.
\cref{eq:positive_score} can be reached because $(\bv{w}_{y} -
\bv{w}_{\hat{y}})\bv{W}_{i}^{\bv{x} - t(\bv{x} - \bv{x}')}(\bv{x} - \bv{x}')$
is the actually classification score $u(\bv{x}), u(\bv{x}')$ difference
between $\bv{x}, \bv{x}'$, and by assumptions \cref{asp:monotony}, they are positive throughout. \cref{eq:segment_integral} is reached due to \cref{cl:1} --- in
this case, the matrix $\bv{W}_l$ in \cref{cl:1} is of rank one.

To arrive from \cref{eq:segment_integral} to
\cref{eq:contraction_lower_bound}, we observe that $\bv{x}'$ is the closest
point to $\bv{x}$ on the decision boundary. Being the
closest means $\bv{x} - \bv{x}' \perp \ca{N}((\bv{w}_y - \bv{w}_{\hat{y}})T')$. If the difference $\bv{x}' -
\bv{x}$ satisfies $\bv{x} - \bv{x}' \not\perp \ca{N}(T')$, we can always
remove the part in the $\ca{N}(T')$, which would identify a point that is
closer to $\bv{x}$, but still on the decision boundary, which would be a
contradiction. Then if $\bv{x} - \bv{x}'$ is orthogonal to the null space, we
can bound the norm using the least singular values. We develop the informal
reasoning above formally in the following.

Similarly in lemma 3.4 in \citet{Jia2019}, by Cauchy interlacing law by row
deletion, assuming $\bv{x} \perp \ca{N}(\prod_{i=1}^{L-1}\bv{W}_i^{q_{j}})$
($\ca{N}$ denotes the null space; the math statement means $\bv{x}$ is
orthogonal to the null space of $\bv{J}(\bv{x})$), we have
\begin{equation}
  ||\prod_{i=1}^{L-1}\bv{W}_{i}^{q_j}\bv{x}||_2 \geq \prod_{i=1}^{L-1}\sigma_{\min}^{i}||\bv{x}||_2
  \label{eq:least_singular_value}
\end{equation}
where $\sigma_{\min}^{i}$ is the smallest singular value of $\bv{W}_i$. Then
conclusion holds as well for multiplication of matrices
$\prod_{i=1}^{L-1}\bv{W}_{i}^{q_j}$, since the multiplication of matrices are
also a matrix.

Notice that in each integral in \cref{eq:segment_integral}, we are integrating over constant. Thus, we have it equates to
\begin{displaymath}
  \sum_{j=1}^{n}(e_j - s_j)\left|(\bv{w}_{y} - \bv{w}_{\hat{y}})^{T}\prod_{i=1}^{L-1}\bv{W}_{i}^{q_j}(\bv{x} - \bv{x}')\right|.
\end{displaymath}

Now we show that in each operand, $\bv{x} - \bv{x}' \perp \ca{N}((\bv{w}_{y} -
\bv{w}_{\hat{y}})^{T}\prod_{i=1}^{L-1}\bv{W}_{i}^{q_j})$.  Denote $T_{q_j}$ as
$\ca{N}((\bv{w}_{y} - \bv{w}_{\hat{y}})^{T}\prod_{i=1}^{L-1}\bv{W}_{i}^{q_j})$.
Suppose that it does not hold. Then we can decompose $\bv{x} - \bv{x}'$ into
two components $\bv{\Delta}_{1}, \bv{\Delta}_{2}$, where $\bv{\Delta}_1 \perp T_{q_j}, \bv{\Delta}_{2}
\not\perp T_{q_j}$. We can find a new point $\bv{x}'' = \bv{x} + \bv{\Delta}_1$ that
is on the boundary. However, in this case
\begin{displaymath}
||\bv{x} - \bv{x}''||_2 = ||\bv{\Delta}_1||_2 \leq ||\bv{\Delta}_1||_2 + ||\bv{\Delta}_2||_2 =
||\bv{x} - \bv{x}'||_2
\end{displaymath}
Recall that $\bv{x}'$ is the closest point to $\bv{x}$ on the decision
boundary. This leads to a contradiction. Repeat this argument for all $j=1,\ldots,
n$, then we have $\bv{x} - \bv{x}'$ be orthogonal to all $\ca{N}(T_{q_j})$.
Thus, by the inequality \cref{eq:least_singular_value} earlier, we can arrive
at \cref{eq:contraction_lower_bound} --- notice that $\bv{w}_y - \bv{w}_{\hat{y}}$ is a
matrix with one column, thus also satisfies the above reasoning.

Through the above inequality, we can transfer the margin to
margin in the instance space. Let $v(\bv{x})$ be the shortest distance in
$||\cdot||_2$ norm from an element $\bv{x} \in \ca{X}$ to the decision
boundary. For a covering ball $C_{i}$, let $v^{i}_{\min}$ be $\min_{\bv{x} \in
  C_i}v(\bv{x})$. Let $v_{\min}$ be the smallest $v^{i}_{\min}$ among all
covering balls
$C_i$ that contain at least a training example. We have that
\begin{displaymath}
  u_{\min} \geq \min_{y, \hat{y} \in \ca{Y}, y\not=\hat{y}}||\bv{w}_{y} - \bv{w}_{\hat{y}}||_2\prod_{i=1}^{L-1}\sigma_{\min}^{i}v_{\min}
\end{displaymath}

Consequently, we can obtain an upper bound of \cref{eq:l_right_case}
parameterized on $v_{\min}$, as follows
\begin{align}\nonumber
  \max\{0, &1 - \frac{u_{\min}}{\gamma}\} \le \max\{0, \\ \nonumber
  & 1 - \frac{\min_{y, \hat{y} \in \ca{Y}, y\not=\hat{y}}||\bv{w}_{y} - \bv{w}_{\hat{y}}||_2\prod_{i=1}^{L-1}\sigma_{\min}^{i}v_{\min}}{\gamma}\}.
\end{align}

Notice that only because $\epsilon_0$-adversarial robustness, we can guarantee that
$v_{\min}$ is non-zero, thus the bound is influenced by AR.

Then, we study case 2), in which all training samples $z \in C_i$ are classified
wrong. In this case, for all $z \in C_i$, the $\hat{y}$ given by $\hat{y}
= \argmax_{i\not= y}\bv{w}^{T}_{i}T'\bv{x}$ in the margin operator is the
same, for that $\hat{y}$ is the wrongly classified class. Its ramp loss is
\begin{displaymath}
  l_{\gamma}(\bv{x}, y) = \max \{0, 1 - \frac{1}{\gamma} (\bv{w}_{y} - \bv{w}_{\hat{y}})^{T}T'\bv{x}\}.
\end{displaymath}
Note that in the case 1), it is the $y$ that stays fixed, while $\hat{y}$ may
differ from example to example; while in the case 2), it is the $\hat{y}$ stays
fixed, while $y$ may differ.

Similarly, within $C_i$ as required by
adversarial robustness, $(\bv{w}_{y} -
\bv{w}_{\hat{y}})^{T}T'\bv{x} \leq 0$, thus we always have $1 -
\frac{1}{\gamma} (\bv{w}_{y} - \bv{w}_{\hat{y}})^{T}T'\bv{x} \geq 1$, implying
\begin{displaymath}
  l_{\gamma}(\bv{x}, y) = 1.
\end{displaymath}
Thus, $\forall z = (\bv{x}, y), z' = (\bv{x}', y') \in C_i$
\begin{equation}
  \label{eq:l_wrong_case}
  |l_{\gamma}(\bv{x}, y) - l_{\gamma}(\bv{x}', y')| = 0.
\end{equation}
Since only these two cases are possible, by \cref{eq:l_right_case} and
\cref{eq:l_wrong_case}, we have $\forall z, z' \in C_i$
\begin{equation}
\label{eq:l_all}
  |l_{\gamma}(\bv{x}, y) - l_{\gamma}(\bv{x}', y')| \leq \max\{0, 1 - \frac{u_{\min}}{\gamma}\}.
\end{equation}

The rest follows the standard proof in algorithmic robust framework.

Let $N_i$ be the set of index of points of examples that fall into $C_i$. Note that $(|N_i|)_{i=1\ldots K}$ is an IDD multimonial random variable with parameters $m$ and $(|\mu(C_i)|)_{i=1\ldots K}$. Then
\begin{align}
        & |R(l \circ T) - R_m(l \circ T)| \nonumber                                                                \\
=    & |\sum\limits_{i=1}^{K}\mathbb{E}_{Z \sim \mu}[l(TX, Y)]\mu(C_i) -
    \frac{1}{m}\sum\limits_{i=1}^{m}l(T\bvec{x}_i, y_i)| \nonumber                                              \\
\leq & |\sum\limits_{i=1}^{K}\mathbb{E}_{Z \sim \mu}[l(TX, Y)]\frac{|N_i|}{m} -
    \frac{1}{m}\sum\limits_{i=1}^{m}l(T\bvec{x}_i, y_i)| \nonumber                                              \\
        & +  |\sum\limits_{i=1}^{K}\mathbb{E}_{Z \sim \mu}[l(TX, Y)]\mu(C_i) -
    \sum\limits_{i=1}^{K}\mathbb{E}_{Z \sim \mu}[l(TX, Y)]\frac{|N_i|}{m}|
    \nonumber                                                                                                   \\
\leq & |\frac{1}{m}\sum\limits_{i=1}^{K}\sum\limits_{j\in N_i}\max\limits_{z \in
    C_i}|l(T\bvec{x}, y) - l(T\bvec{x}_j, y_j)| \label{t:1} \\
        & + |\max\limits_{z \in \mathcal{Z}}|l(T\bvec{x},
    y)|\sum\limits_{i=1}^{K}|\frac{|N_i|}{m} - \mu(C_i)|| \label{t:2}.
\end{align}
Remember that $z = (\bvec{x}, y)$.

By \cref{eq:l_all} we have \cref{t:1} is equal or less than $\max\{0, 2(1 - \frac{u_{\min}}{\gamma})\}$.
By Breteganolle-Huber-Carol inequality, \cref{t:2} is less or equal to
$\sqrt{\frac{\log(2)2^{k+1}C_{\mathcal{X}}^{k}}{\gamma^{k}m} + \frac{2
    \log(1/\eta)}{m}}$.

The proof is finished.
\end{proof}

%\clearpage
\pagebreak
%-------------------------------------------------------------------------
\section{Implementation Details}
\label{sec:exps}
We summarize the details of the experiments in this section. The experiments
are run with PyTorch \citep{Pfeiffer2007}.

\begin{figure}[h]
  \centering
    \includegraphics[width=\columnwidth]{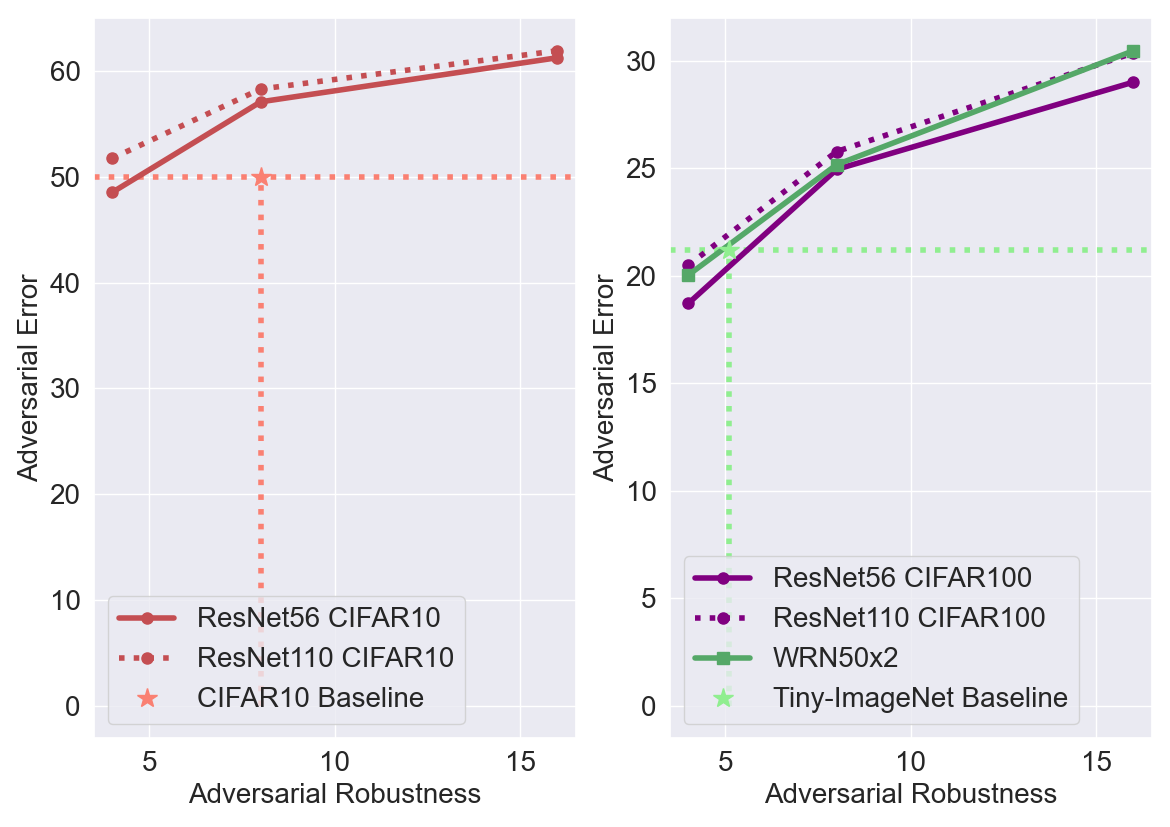}
  \caption{The plot of accuracy on {\it adversarial examples}
v.s. adversarial defense strength built in NNs. The dotted line of which the
intersections are marked by stars are adversarial accuracy in \citet{Madry2017}
(CIFAR10), in \citet{Li2018a} (Tiny ImageNet) under
similar adversarial attack strength.}
  \label{fig:defense}
\end{figure}

\subsection{Datasets}
\noindent{\bf{CIFAR10/100}}.
Each CIFAR dataset consists of $50,000$ training data and $10,000$ test data.
CIFAR-10 and CIFAR-100 have $10$ and $100$ classes respectively.  Our data
augmentation follows the standard manner in \citet{Lee2015}: during training, we
zero-pad $4$ pixels along each image side, and sample a $32\times 32$ region
cropped from the padded image or its horizontal flip; during testing, we use the
original non-padded image.

\noindent{\bf{Tiny-ImageNet}}.  Tiny-ImageNet is a subset of ImageNet dataset,
which contains $200$ classes rather than $1,000$ classes.  Each class has $500$
training images and $50$ validation images.  Images in the Tiny-ImageNet
dataset are of $64\times64$ pixels, as opposed to $256\times256$ in the full
ImageNet set.  The data augmentation is straightforward: an input image is
$56\times56$ randomly cropped from a resized image using the scale, aspect
ratio augmentation as well as scale jittering. A single $56\times56$
cropped image is used for testing.

\subsection{Experiments in \cref{sec:regul-effects-nns}
}
\label{sec:exp_1}

\begin{table}[t]
  \caption{
    Raw data of CIFAR10 dataset for plots in \cref{fig:datasets}, \cref{fig:datasets_resnet110} and \cref{fig:defense}.
}
	\label{table:Relaxation_constraint_CIFAR10}
	\vskip 0.15in
	\begin{center}
		\begin{small}
			%% \subcaption{
				\begin{tabular}{cc c c c}
					\toprule
					\multirow{2}{*}{\makecell{Method}} & \multirow{2}{*}{ } & \multicolumn{3}{c}{Defensive Strength} \\
					\cline{3-5}
					~ & ~ & 4 & 8 & 16\\
					\midrule
					\multirow{7}{*}{\makecell{ResNet-56 \\ + Adv Trn}}
					~ & Trn Acc. & $99.51$ & $98.45$ & $95.97$ \\
					~ & Test Acc. & $88.86$ & $87.51$ & $84.89$ \\
					~ & $\Delta$Acc. & $10.65$ & $10.94$ & $11.08$ \\
					\cline{2-5}
					~ & Trn Loss & $0.014$ & $0.043$ & $0.105$ \\
					~ & Test Loss & $0.683$ & $0.649$ & $0.650$ \\
					~ &	$\Delta$Loss & $0.669$ & $0.606$ & $0.545$   \\
					\cline{2-5}
					~ & PGD & $65.92$ & $65.24$ & $72.16$ \\
					\midrule
					\multirow{7}{*}{\makecell{ResNet-110 \\ + Adv Trn}}
					~ & Trn Acc. & $99.95$ & $99.62$ & $98.42$ \\
					~ & Test Acc. & $89.20$ & $87.09$ & $85.02$ \\
					~ & $\Delta$Acc. & $10.75$ & $12.53$ & $13.40$ \\
					\cline{2-5}
					~ & Trn Loss & $0.002$ & $0.010$ & $0.044$ \\
					~ & Test Loss & $0.825$ & $0.813$ & $0.729$ \\
					~ &	$\Delta$Loss & $0.823$ & $0.803$ & $0.685$  \\
					\cline{2-5}
					~ & PGD & $58.02$ & $66.94$ & $72.40$\\
					\bottomrule
				\end{tabular}
			%% }
		\end{small}
	\end{center}
	\vskip -0.1in
\end{table}

\noindent{\bf{CIFAR10/100} Models and Training}.
The models for CIFAR10/100 are the same as the ones in
\cref{sec:regul-effects-nns-1}, except that we do not use spectral
normalization anymore. CIFAR100 has 100 output neurons instead of 10.

\noindent{\bf{Tiny-ImageNet Model}}.
For Tiny ImageNet dataset, we use $50$-layered wide residual networks with $4$
groups of residual layers and $[3, 4, 6, 3]$ bottleneck residual units for each
group respectively.  The $3\times3$ filter of the bottleneck residual units
have $[64\times k, 128\times k, 256\times k, 512\times k]$ feature maps with
the widen factor $k = 2$ as mentioned in \citet{Zagoruyko2016}.  We replace the
first $7\times7$ convolution layer with $3\times3$ filters with stride $1$ and
padding $1$.  The max pooling layer after the first convolutional layer is also
removed to fit the $56\times56$ input size.  Batch normalization layers are
retained for this dataset.  The weights of convolution layers for Tiny ImageNet
are initialized with Xavier uniform \citep{Glorot2010}.  Again, all dropout
layers are omitted.

\noindent{\bf{Tiny-ImageNet Training}}.
The experiments on the Tiny-ImageNet dataset are based on a mini-batch size of
$256$ for $90$ epochs. The initial learning rate is set to be $0.1$ and decayed at
$10$ at $30$ and $60$ epochs respectively.  All experiments are trained on the
training set with stochastic gradient descent with the momentum of $0.9$.

\noindent{\bf{Results}}.
The data for \cref{fig:datasets} and \cref{fig:datasets_resnet110} are given in
\cref{table:Relaxation_constraint_CIFAR10},
\cref{table:Relaxation_constraint_CIFAR100} and \cref{table:Exp_Tiny-ImageNet}.
More specifically, the data on CIFAR10 are given in
\cref{table:Relaxation_constraint_CIFAR10}.  The result on CIFAR100 are given
in \cref{table:Relaxation_constraint_CIFAR100}.  The result on Tiny-ImageNet
are given in \cref{table:Exp_Tiny-ImageNet}.

\noindent{\bf{Adversarial Robustness Attack Method.}}
The adversarial accuracy is evaluated against $l_{\infty}$-PGD \citep{Madry2017}
untargeted attack adversary, which is one of the strongest white-box attack
methods. When considering adversarial attack, they usually train and evaluate
against the same perturbation. And for our tasks, we only use the moderate
adversaries that generated by $10$ iterations with steps of size $2$ and
maximum of $8$. When evaluating adversarial robustness, we only consider clean
examples classified correctly originally, and calculate the accuracy of the
adversarial examples generated from them that are still correctly
classified. The adversarial accuracy is given in
\cref{table:Relaxation_constraint_CIFAR10}
\cref{table:Relaxation_constraint_CIFAR100} \cref{table:Exp_Tiny-ImageNet}, the
row named ``PGD'', and plotted in \cref{fig:defense}.

\begin{table}[t]
  \caption{
    Raw data of CIFAR100 dataset for the plot in \cref{fig:datasets}, \cref{fig:datasets_resnet110}
    and \cref{fig:defense}.
  }
	\label{table:Relaxation_constraint_CIFAR100}
	\vskip 0.15in
	\begin{center}
		\begin{small}
			%% \subcaption{
				\begin{tabular}{ccc c c}
					\toprule
					\multirow{2}{*}{Method} & \multirow{2}{*}{ } & \multicolumn{3}{c}{\makecell{Defensive Strength}} \\
					\cline{3-5}
					~ & ~ & 4 & 8 & 16\\
					\midrule
					\multirow{7}{*}{\makecell{ResNet-56 \\ + Adv Trn}}
					& Trn Acc. & $88.73$ & $86.97$ & $82.17$ \\
					& Test Acc. & $61.31$ & $60.87$ & $59.43$ \\
					~ & $\Delta$Acc. & $27.42$ & $26.10$ & $22.74$ \\
					\cline{2-5}
					~ & Trn Loss & $0.357$ & $0.413$ & $0.570$ \\
					~ & Test Loss & $2.063$ & $2.106$ & $1.978$ \\
					& $\Delta$Loss   & $1.706$ & $1.693$ & $1.408$ \\
					\cline{2-5}
					~ & PGD   & $30.52$ & $40.99$ & $48.81$ \\
					\midrule
					\multirow{7}{*}{\makecell{ResNet-110 \\ + Adv Trn}}
					& Trn Acc. & $96.91$ & $94.55$ & $90.90$ \\
					& Test Acc. & $61.48$ & $61.26$ & $59.56$ \\
					~ & $\Delta$Acc. & $35.43$ & $33.29$ & $31.34$ \\
					\cline{2-5}
					~ & Trn Loss & $0.098$ & $0.171$ & $0.278$ \\
					~ & Test Loss & $2.645$ & $2.413$ & $2.323$ \\
					& $\Delta$Loss   & $2.547$ & $2.241$ & $2.045$ \\
					\cline{2-5}
					~ & PGD   & $33.33$ & $42.08$ & $50.99$ \\
					\bottomrule
				\end{tabular}
			%% }
		\end{small}
	\end{center}
	\vskip -0.1in
\end{table}

\begin{table}[t]
  \caption{
    Raw data of Tiny-ImageNet dataset for the plot in \cref{fig:datasets}, \cref{fig:datasets_resnet110}
    and \cref{fig:defense}.
  }
	\label{table:Exp_Tiny-ImageNet}
	\vskip 0.15in
	\begin{center}
		\begin{small}
			%% \subcaption{
				\begin{tabular}{cc c c c c}
					\toprule
					\multirow{2}{*}{Method} & \multirow{2}{*}{} & \multicolumn{4}{c}{\makecell{Defensive Strength}} \\
					\cline{3-6}
					~ & ~ & 0 & 4 & 8 & 16\\
					\midrule
					\multirow{7}{*}{\makecell{Wide ResNet
                                  \\ + Adv Trn}}
					~ & Trn Acc. & $79.12$ & $73.71$ & $66.17$ & $60.73$ \\
					~ & Test Acc. & $63.43$ & $62.09$ & $61.09$ & $57.36$ \\
					~ & $\Delta$Acc. & $15.69$ & $11.62$ & $5.08$ & $3.37$ \\
					\cline{2-6}
					~ & Trn Loss & $0.874$ & $1.080$ & $1.384$ & $1.641$ \\
					~ & Test Loss & $1.561$ & $1.637$ & $1.689$ & $1.806$ \\
					~ &	$\Delta$Loss &$0.687$ & $0.557$ & $0.305$ & $0.165$ \\
					\cline{2-6}
					~ & PGD  & $0.00$ & $32.26$ & $41.20$ & $53.12$ \\
					\bottomrule
				\end{tabular}
			%% }
		\end{small}
	\end{center}
	\vskip -0.1in
\end{table}

\subsection{Experiments in \cref{sec:regul-effects-nns-1}}

\begin{table}[t]
  \caption{
    Raw data for \cref{fig:control}.
    SP denotes spectral norm.
  }
	\label{table:ResNet56_Gradually_Relaxation_constraint_CIFAR10}
	\vskip 0.15in
	\begin{center}
		\begin{small}
			%% \subcaption{
				\begin{tabular}{cc c c c}
					\toprule
					\multirow{2}{*}{\makecell{Strength of \\ Spectral Normalization}} & \multirow{2}{*}{ } & \multicolumn{3}{c}{Defensive Strength} \\
					\cline{3-5}
					~ & ~ & 4 & 8 & 16\\
					\midrule
					\multirow{7}{*}{\makecell{SP 1}}
					~ & Trn Acc. & $96.91$ & $94.38$ & $90.58$ \\
					~ & Test Acc. & $90.47$ & $88.87$ & $85.82$ \\
					~ & $\Delta$Acc. & $6.44$ & $5.51$ & $4.76$\\
					\cline{2-5}
					~ & Trn Loss & $0.092$ & $0.159$ & $0.265$ \\
					~ & Test Loss & $0.316$ & $0.353$ & $0.432$ \\
					~ &	$\Delta$Loss & $0.224$ & $0.194$ & $0.168$ \\
					\cline{2-5}
					~ & PGD  & $57.93$ & $69.98$ & $75.98$  \\
					\midrule
					\multirow{7}{*}{\makecell{SP 3}}
					~ & Trn Acc. & $99.65$ & $98.51$ & $95.94$ \\
					~ & Test Acc. & $90.02$ & $88.07$ & $85.43$ \\
					~ & $\Delta$Acc. & $9.63$ & $10.44$ & $10.51$ \\
					\cline{2-5}
					~ & Trn Loss & $0.010$ & $0.039$ & $0.107$ \\
					~ & Test Loss & $0.606$ & $0.580$ & $0.577$  \\
					~ &	$\Delta$Loss & $0.596$ & $0.541$ & $0.470$ \\
					\cline{2-5}
					~ & PGD  & $56.83$ & $67.73$ & $73.41$ \\
					\midrule
					\multirow{7}{*}{\makecell{SP 5}}
					~ & Trn Acc.& $99.57$ & $98.33$ & $95.96$ \\
					~ & Test Acc. & $89.53$ & $88.09$ & $85.32$ \\
					~ & $\Delta$Acc. & $10.04$ & $10.24$ & $10.64$ \\
					\cline{2-5}
					~ & Trn Loss & $0.012$ & $0.045$ & $0.105$ \\
					~ & Test Loss & $0.649$ & $0.602$ & $0.611$  \\
					~ &	$\Delta$Loss & $0.638$ & $0.557$ & $0.506$ \\
					\cline{2-5}
					~ & PGD  & $54.91$ & $65.96$ & $72.37$ \\
					\midrule
					\multirow{7}{*}{\makecell{SP Uncontrolled}}
					~ & Trn Acc. & $99.51$ & $98.45$ & $95.97$ \\
					~ & Test Acc. & $88.86$ & $87.51$ & $84.89$ \\
					~ & $\Delta$Acc. & $10.65$ & $10.94$ & $11.08$ \\
					\cline{2-5}
					~ & Trn Loss & $0.014$ & $0.043$ & $0.105$ \\
					~ & Test Loss & $0.683$ & $0.649$ & $0.650$ \\
					~ &	$\Delta$Loss & $0.669$ & $0.606$ & $0.545$ \\
					\cline{2-5}
					~ & PGD & $65.92$ & $65.24$ & $72.16$\\
					\bottomrule
				\end{tabular}
			%% }
		\end{small}
	\end{center}
	\vskip -0.1in
\end{table}

\noindent{\bf{Models}}.
We use ResNet-type networks \citep{Zhang2019a}. Given
that we need to isolate factors that influence spectral complexity, we use ResNet without
additional batch normalization (BN) layers. To train ResNet without BN, we rely
on the fixup initialization proposed in \citet{Zhang2019a}. The scalar layers in
\citet{Zhang2019a} are also omitted, since it changes spectral norms of
layers. Dropout layers are omitted as well. Following \citet{Sedghi2018a}, we
clip the spectral norm every epoch rather than every iteration.

\noindent{\bf{Training}}.
The experiments on CIFAR10 datasets are based on a mini-batch size of $256$ for
$200$ epochs.  The learning rate starts at $0.05$, and is divided by $10$ at
$100$ and $150$ epochs respectively.  All experiments are trained on training
set with stochastic gradient descent based on the momentum of $0.9$.

\noindent{\bf{Results}}.
The data for \cref{fig:control} are given in \cref{table:ResNet56_Gradually_Relaxation_constraint_CIFAR10}.

\end{document}